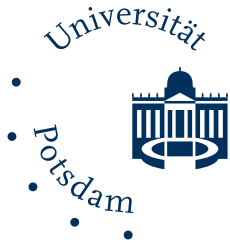
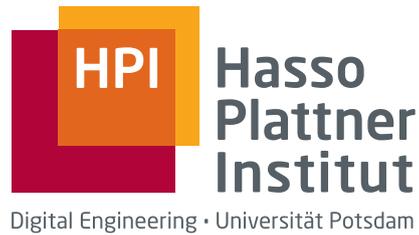

# Hallucination Detection with the Internal Layers of LLMs

Halluzinationserkennung mittels internerer Schichten großer Sprachmodelle

## Martin Preiß

Universitätsmasterarbeit
zur Erlangung des akademischen Grades

## Master of Science
*(M. Sc.)*

im Studiengang
IT Systems Engineering

eingereicht am 10. Juni 2025 am
Fachgebiet Algorithm Engineering der
Digital-Engineering-Fakultät
der Universität Potsdam

| | |
|---|---|
| **Gutachter** | Prof. Dr. Gerard de Melo |
| | Prof. Dr. Christoph Lippert |
| **Betreuer** | Yindong Wang |

# Abstract

Large Language Models (LLMs) have succeeded in a variety of natural language processing tasks [Zha+25]. However, they have notable limitations. LLMs tend to generate hallucinations, a seemingly plausible yet factually unsupported output [Hua+24], which have serious real-world consequences[Kay23; Rum+24]. Recent work has shown that probing-based classifiers that utlize LLMs' internal representations can detect hallucinations [AM23; Bei+24; Bur+24; DYT24; Ji+24; SMZ24; Su+24]. This approach, since it does not involve model training, can enhance reliability without significantly increasing computational costs.

Building upon this approach, this thesis proposed novel methods for hallucination detection using LLM internal representations and evaluated them across three benchmarks: TruthfulQA, HaluEval, and ReFact. Specifically, a new architecture that dynamically weights and combines internal LLM layers was developed to improve hallucination detection performance. Throughout extensive experiments, two key findings were obtained: First, the proposed approach was shown to achieve superior performance compared to traditional probing methods, though generalization across benchmarks and LLMs remains challenging. Second, these generalization limitations were demonstrated to be mitigated through cross-benchmark training and parameter freezing. While not consistently improving, both techniques yielded better performance on individual benchmarks and reduced performance degradation when transferred to other benchmarks. These findings open new avenues for improving LLM reliability through internal representation analysis.

# Zusammenfassung

Große Sprachmodelle (Large Language Models, LLMs) haben sich bei zahlreichen Aufgaben der Verarbeitung natürlicher Sprache bewährt [Zha+25]. Ihre Auswirkungen auf die Gesellschaft sind jedoch nicht nur positiv. LLMs neigen dazu, Halluzinationen zu erzeugen, eine scheinbar plausible, aber faktisch nicht belegte Ausgabe [Hua+24], die der Gesellschaft nachweislich schadet[Kay23; Rum+24]. Jüngste Forschungen haben gezeigt, dass Klassifikatoren (genannt Probes) die die internen Schichten von LLMs nutzen, für die Erkennung von Halluzinationen genutzt werden können [AM23; Bei+24; Bur+24; DYT24; Ji+24; SMZ24; Su+24]. Da die LLMs nicht neu trainiert werden müssen, könnte diese Methode die Zuverlässigkeit steigern, ohne den Rechenaufwand wesentlich zu erhöhen.

Aufbauend auf diesem Ansatz werden in dieser Arbeit neue Methoden zur Erkennung von Halluzinationen unter Verwendung interner LLM-Repräsentationen präsentiert und anhand von drei Datensätzen evaluiert: TruthfulQA, HaluEval und ReFact. Insbesondere wurde eine neue Architektur entwickelt, welche die internen LLM-Schichten dynamisch gewichtet und kombiniert, um die Leistung von Halluzinationserkennungen zu verbessern. Im Rahmen umfangreicher Experimente wurden zwei wichtige Erkenntnisse gewonnen: Erstens konnte gezeigt werden, dass der vorgeschlagene Ansatz im Vergleich zu traditionellen Methoden zur Erkennung von Halluzinationen eine bessere Leistung erzielt, auch wenn die Generalisierung über Datensätze und LLMs hinweg herausfordernd bleibt. Zweitens wurde gezeigt, dass diese Generalisierungseinschränkungen durch datensatzübergreifendes Training und Einfrieren der Gewichte gemildert werden können. Obwohl beide Techniken keine konsistente Verbesserung brachten, erzielten sie eine bessere Leistung bei einzelnen Benchmarks und verringerten den Leistungsabfall bei der Übertragung auf andere Benchmarks. Diese und andere gewonnene Erkenntnisse eröffnen neue Möglichkeiten zur Verbesserung der LLM-Zuverlässigkeit durch interne Repräsentationsanalyse.

# Acknowledgments

Most of the people that I want to acknowledge are of german mother tongue. Therefore, this acknowledgment is written in german. Still, an English version can be found below.

**German Version**

Mit diesem 82 Seiten langem Stapel an Papier geht dann wohl nun meine Zeit als Student zu Ende. Ohne die Unterstützung meiner Familie wäre das unmöglich gewesen und ein spezieller Dank geht deshalb an sie. Vor allem an meine Eltern, die mich finanziell gefördert und bei den 3 Umzügen der letzte 10 Monate geholfen habe. Ein spezieller Dank geht auch an alle Freunde hier in Potsdam und im Rest der Welt, die die stressigsten Phasen immer sehr ertragbar werden ließen. In Bezug auf die Thesis müssen besonders Moritz und Nick hervorgehoben werden, ohne die ich wahrscheinlich dem Wahnsinn verfallen wär, hätten sie nicht stets neben mir an Ihrer Thesis gearbeitet. Aber auch Leon und Cedric darf ich nicht vergessen, die es mir durch ihre Untermiete überhaupt ermöglicht haben weiter in Potsdam zu leben.

Nicht zu vergessen sind natürlich alle Mitglieder des Masterprojekts, die in unerlässlichem Eifer 1001 Datenpunkt von ReFact kontrolliert haben. Spezieller Dank, geht auch an Yindong deren Feedback die Qualität dieser Arbeit wohl um einiges angehoben hat. Auch muss ich dem De Melo Lehrstuhl sowie dem KISZ für die Zurverfügungstellung der Ressourcen danken. Natürlich, geht auch ein Dank an die beiden Professoren, Prof. Dr.de Melo und Prof. Dr. Lippert, die sich die Freude herausnehmen diese Arbeit zu kontrollieren und über die letzten 6 Jahre zu viel meiner Wasseranreicherung beigetragen haben. Dieser Dank richtet sich auch speziell an alle Mitarbeiter des HPIs. Vor allem auch an alle Servicekräfte der Mensa.

Zuletzt möchte ich an dieser Stelle noch den Glückskeks zitieren, der mir vor zwei Monaten mitgeteilt hat, ich solle doch nicht rumheulen:

*"Widerstände spornen sie an! Also ranklotzen und lospoweren"*

Viel Spaß beim Lesen :)

**English Version**

With this 82-page pile of paper, my time as a student has come to an end. Without the support of my family this would have been impossible and a special thanks goes to them. Especially to my parents who supported me financially and helped me with the 3 relocations of the last 10 months. Special thanks also go to all my friends here in Potsdam and in the rest of the world, who always made the most stressful phases very bearable. With regard to the thesis, special acknowledgement must go to Moritz and Nick, without whom I would probably have gone crazy if they hadn't always worked alongside me on their thesis. But Leon and Cedric must also be emphasized, who made it possible for me to continue living in Potsdam due to subletting them me their flat.

Not to forget, the members of the master project needed to by acknowledge which annotated 1001 data points of ReFact. Special thanks also goes to Yindong whose feedback has probably raised the quality of this work a lot. I must also thank the De Melo Chair and the KISZ for providing the resources. Of course, thanks also go to the two professors, Prof. Dr. de Melo and Prof. Dr. Lippert, who have to control this work and have contributed to the enrichment of my knowledge over the last 6 years. I would also like to express my special thanks to all HPI employees. Especially to all the service staff in the canteen.

Finally, I would like to quote the fortune cookie that told me two months ago not to complain:

*"Resistance motivates you! So work hard and get going."*

Have fun reading :)

# Contents

# 1 Introduction

## 1.1 Motivation

Since the release of ChatGPT, Large Language Models (LLMs) have succeeded in natural language processing [Zha+25]. By leveraging billions of parameters trained on large textual corpora, they demonstrated capabilities to understand and generate textual content in a wide range of user tasks. Consequently, LLMs continue to have a significant impact on society, improving the process of information retrieval and processing.

However, the impact is not solely positive, as LLMs tend to generate hallucinations, a seemingly plausible yet factually unsupported output [Hua+24]. One example is the case of Brian Hood, an Australian major who ChatGPT falsely claimed to have served time in prison for bribery [Kay23]. This resulted in the first defamation lawsuit against OpenAI regarding generated content. Another crucial example is the work of [Rum+24] showing the tendency of LLMs to hallucinate when asked to summarize health records. With that, LLMs could mislead the diagnostic process and lead to the direct harm of human beings.

The master project "Facts Matter" 2024 of the De Melo Chair at HPI examined this in depth [Wan+25]. By creating a new benchmark called ReFact via hidden fake facts, multiple state-of-the-art LLMs were evaluated in detecting and locating factually incorrect information. The project's results highlighted the unreliability of LLMs in handling fine-grained fake facts due to hallucinations.

This lack of trust and the potential risk of inducing fake facts into society motivate effective and practicable hallucination detection methods. Although traditional approaches can perform well, they are not without limitations [Hua+24]. For example, approaches that leverage uncertainty express overconfidence, while fact-checking strategies require reliable knowledge sources.

Recent work showed that the layers of LLMs, internal architectural components, encode representations closely related to hallucinations [Ahd+24; AM23; Bur+24; CRG23; Li+24b; MT24]. By building small classifiers (probes) on the internal layers of LLMs, researchers found representations of existing knowledge, truth or lying behavior. Extending this work, papers were published that demonstrate the potential of these models to detect hallucinations [Bei+24; DYT24; Ji+24; SMZ24;

Su+24]. As no retraining of the LLMs is required, this approach can enhance the reliability without significantly increasing the computational costs.

Still, these methods are not without critique. For example, the study of [LH24] argues that current probing techniques can't succeed for conceptual reasons and concludes that they fail to generalize adequately.

However, as this domain is relatively new, there is room for improvement. The work of [Bei+24; CH-+24; Ji+24; SMZ24] revealed patterns suggesting that important information is encoded in the middle to latter layers. One popular example, which also exploits this, is the decoding strategy DoLa [Chu+24]. By incorporating the latter layers into the decoding process, DoLa improves the truthfulness of LLMs. This raises the question whether a method that adjusts the impact of different layers can reduce the generalization issues and improve the overall performance of hallucination detectors.

In addition, the unique characteristics of the ReFact benchmark reveal new possibilities in this research domain [Wan+25]. As the provided samples are relatively long and contain hidden fake facts, evaluating probing-based methods with ReFact might show whether they are capable of performing under an increased complexity. Furthermore, spans are provided for every sample, enabling experiments regarding the identification of factually incorrect data inside the text.

## 1.2  Contribution

This thesis continues to explore the potential of detecting hallucinations with the internal layers of LLMs on the benchmarks ReFact[Wan+25], HaluEval[Li+23], and TruthfulQA[LHE22b]. Specifically, a new architecture was derived that leverages these internal layers as an input to classify hallucinated outputs of LLMs in a fake fact detection task. Afterward, its capabilities were evaluated in generalization tests, pretraining/freezing evaluations, and baseline comparisons. Additionally, the performance of the new model in sequential labeling was investigated.

Consequently, the first research question **RQ1** investigates how a method that takes advantage of LLM layers can be built. Although related work provides potential model designs, explored in section 3.2, this work proposes a new architecture. The reports of layer importance by [Bei+24; CH-+24; Ji+24; SMZ24] motivated to design a model that adjusts and compares the impact of different layers. In total, 130 different model variations were explored. Experiments showed that models that leverage a shared MLP over every layer and optionally compare the resulting encodings with cosine similarity, output practicable classification features. With that, the proposed methods achieved F1 scores up to 0.82 on the HaluEval benchmark, and two of the models explored were selected to be investigated in the upcoming research questions. The first one utilizes the cosine similarity, while second one doesn't. hroughout this thesis they are stated as *cosine* and *no comparison.*

Due to the earlier-mentioned generalization issues, a major concern of this thesis was to be more skeptical. For this reason, the second research question **RQ2** explores whether the proposed architecture transfers to other benchmarks or other LLMs. The results showed that the developed method can improve a simulated fake fact detection scenario on the test split of their training benchmark. Nevertheless, like the related work, the models did not generalize across benchmarks. Training on 8 different LLMs revealed that the no comparison model stayed constant in performance, while the cosine model decreased with increasing LLM size and label imbalance.

Furthermore, the third research question **RQ3** tested whether successive training on 2 benchmarks can enhance performance or mitigate the generalization issues. As the proposed method learns weights corresponding to LLM layers, tests were conducted to measure how freezing the weights after training on the first benchmark affects the performance. Experimental results indicated that neither idea consistently improved the methodology. However, in particular scenarios, both approaches enabled the models to achieve their maximum measured performance on each benchmark and reduced the decrease when transferred to other benchmarks.

The fourth research question **RQ4** focused on comparing the methodology to the related work. Multiple baselines were implemented that leverage the internal layers of LLMs. While not transferring effectively to unseen benchmarks, the methodology of this thesis outperformed all of the implemented baselines on average. This demonstrates the potential of the developed architecture.

In the end, the fifth research question **RQ5** explored whether the method enables the identification of hallucinatory spans provided by the ReFact benchmark. In total, six different approaches were tested with the baselines and the methodology of this thesis. Despite outperforming the baselines and achieving a high accuracy of up to 96%, the method did not identify the marked hallucination spans precisely, likely due to a high label imbalance.

To summarize, the research questions answered in this thesis are:

- **RQ1**: How can a hallucination detection method be built that leverages the internal layers of LLMs?
- **RQ2**: To what extent does the proposed method generalize across other benchmarks or to other LLMs?
- **RQ3**: How do learned weight distributions corresponding to LLM layers affect the performance of the method combined with pretraining or freezing?
- **RQ4**: How does the method compare to other hallucination detection methods that utilize the internal layers of LLMs?
- **RQ5**: How precisely can the proposed method identify hallucinated spans within the input text?

The remaining content of this thesis is structured as follows. At first, in chapter 2, important concepts to understand this thesis are introduced. Then the related work is described in chapter 3. Afterward, the methodology is presented in chapter 4. In chapter 5, the experiments corresponding to the research questions are described, which are discussed in chapter 6. Finally, the thesis is concluded in chapter 7.

All code used to build and evaluate the method is publicly available at https://github.com/MartinPreiss/MasterThesis.

# 2 Background

In this chapter, important concepts and foundations are introduced. For a more detailed explanation, refer to the cited literature.

## 2.1 Large Language Models

The next section describes the taxonomy of LLMs, their internal architecture, and the different phases of their training and deployment.

### 2.1.1 Taxonomy of Language Models

Language Modeling (LM) aims to model language intelligence for machines and predicts outputs by modeling the generative likelihood of the input sequence [Zha+25]. Early focus in research was on statistical learning. However, this shifted in the direction of neural language models, which represent the probability of word sequences with neural networks, like the later introduced MLP. The first neural networks were inspired by the biological brain and are mathematical functions that hold learnable parameters to perform a prediction [Zha+23a]. The term Large Language Models (LLM) usually refers to transformer-based neural language models [Min+25]. Large indicates the incredible size of their parameters and training data, as they contain multiple billions of parameters, pretrained on massive textual corpora. The word transformer designates the underlying architecture.

### 2.1.2 The Transformer Architecture

First introduced in the paper "Attention is All You Need" [Vas+23] the transformer was responsible for the success of LLMs in recent years[Zha+25]. An in-depth description of the transformer architecture would be beyond scope. Nonetheless, it is summarized shortly using the content of [JM25]. Predicting with a transformer requires splitting the input text, called prompt, into small subwords, called tokens. Tokens can be words or parts of words or even individual letters. This step is called tokenization and is handled by specific algorithms. Afterwards, the tokens are mapped to vocabulary indices and this list of indices are the input of the transformer.

Via a matrix, which holds the initial values, the indices are mapped to vectors called embeddings. A positional encoding is added, so that the information about the input location is kept. The embeddings serve as input to a stack of transformer blocks, mathematical functions that consist of residual connections, normalizing layers, MLPs, and attention layers. These stacks are the layers of LLMs. The output of one layer serves as input for the next layer, and by that, each token embedding is enriched with sequential information of the neighboring tokens, layer by layer.

After passing the embeddings through the transformer blocks, they are usually unembedded with an MLP, resulting in vectors of vocabulary size. The values of those vectors are called logits and, via a mathematical function called the softmax, mapped into a probability space. With that, the transformer outputs the probability of the next token for every sequential input. The task of selecting the final generation is called decoding, which is achieved through specific algorithms that take advantage of the final token distributions. A figure visualizing the transformer is shown in Figure 2.1. The transformer was proposed as an encoder-decoder architecture. However, the current state-of-the-art LLMs are mainly decoder-based, only including previous tokens. Therefore, only decoder-based transformers are discussed in this thesis.

### 2.1.3  Life Cycle of Large Language Models

LLMs are typically autoregressive models, iteratively predicting the next word based on the previous generation. This task of autoregressive word prediction is capable of solving multiple other NLP tasks, like question answering or summarization. To accomplish, an LLM lives through multiple phases called the LLM life cycle (pretraining → alignment → inference) [Zha+23b]. The first phase is the pretraining stage in which LLMs are trained on a large textual corpus to incorporate knowledge into their parameters, often called parametric knowledge. Afterwards, in the alignment phase, they learn to solve downstream tasks via supervised fine-tuning or reinforcement learning from human feedback. The process of fitting language models to follow text instructions is often called instruction-finetuning [JM25]. For the rest of this thesis, if not stated otherwise, the word LLM serves as a reference for a transformer-based instruction fine-tuned decoder.

## 2.2  Hallucinations

This section describes the taxonomy of hallucinations, their sources and benchmarks that enable the development of detection methods.

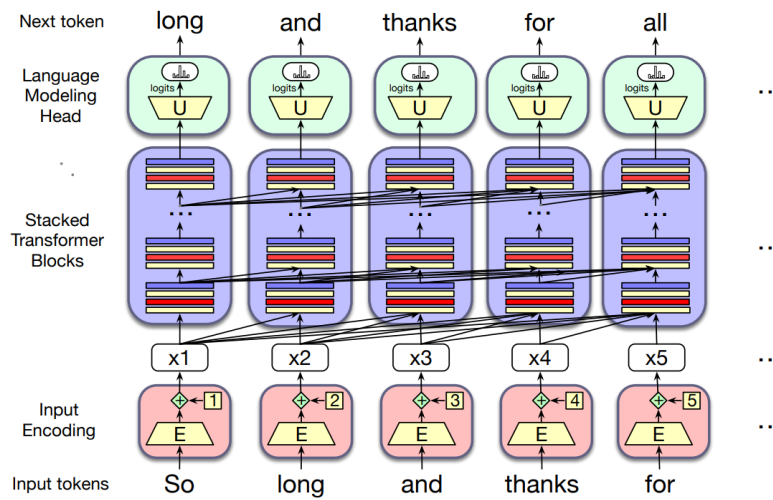

**Figure 2.1:** A visualization of the transformer architecture created by [JM25]

### 2.2.1 Taxonomy of Hallucinations

In Natural Language Processing (NLP) hallucinations are defined as generated content that appears nonsensical or unfaithful to the provided source. [Hua+24; Zha+23b]. For LLMs specifically, this represents a seemingly plausible yet factually unsupported output [Hua+24]. Further, following the taxonomy of [Hua+24], hallucinations are classified into 2 groups, namely factual and faithful hallucinations. Factual hallucinations are defined as generated content that conflicts with verifiable world knowledge. In contrast, faithful hallucinations are defined as generated content that is self-conflicting or in conflict with the provided source. As ReFact's focus is on scientific facts [Wan+25], the scope of this thesis is reduced to factual hallucinations. For this reason, this work only refers to factual hallucinations as hallucinations if not stated otherwise.

### 2.2.2 Sources of Hallucinations

There are multiple reasons why LLM's hallucinate. [Zha+23b] identified 4 primary sources: the parametric memory, overconfidence, misleading alignment, and the generation strategy, which are all introduced at different stages of the LLM life cycle. First, the parametric knowledge of the model can lack relevant information or could have incorporated biased, fabricated or outdated facts due to the training and alignment data, misleading the model. Second, LLMs predict overconfident probabilities. [Hua+24] explains this inability to provide well-calibrated uncertainty scores with the nature of the softmax and other architectural problems of LLMs, like fading out attention weights. Third, the alignment stages might introduce misleading behavior, for example, forcing the model to output non-existent knowledge to favor the user's perspective. Furthermore, the autoregressive approach of LLMs can result in snowballing effects, as the model has to stick to previously made errors. Finally, induced randomness in the decoding strategies can result in incorrect predictions.

### 2.2.3 Hallucination Detection Benchmarks

Multiple methods have been proposed to address these sources [Hua+24; Ji+23; Zha+23b]. However, the problem remains, which is why effective detection methods are crucial. Hallucination benchmarks have been created to evaluate methods that detect and mitigate hallucinations [Hua+24]. These benchmarks are the backbone of this thesis, as they enable a quantitative exploration of hallucinatory patterns. Current benchmarks focus on fact-conflicting hallucinations and evaluate them across various tasks like natural language generation or question answering[Zha+23b]. They are usually created by humans or at least annotated by them, as they are more reliable. However, automatic generation may be the case, with filtering to achieve a certain threshold of quality. Most benchmarks focus either on employing LLMs to generate or to discriminate hallucinations. For the latter, these benchmarks provide truthful and factually wrong responses. As ReFact is a discriminatory factual QA benchmark[Wan+25], this thesis only utilizes this category.

## 2.3 Methodological Foundations

To understand the proposed method, key concepts have to be introduced. At first, the overall problem is formulated. Afterward, important foundations are described, like supervised learning, probes, MLPs, and methods to measure similarities.

### 2.3.1 Problem Formulation

The task of hallucination detection is defined as distinguishing between hallucinated and factual content in a text [Hua+24]. When classifying the whole textual sequence as either containing hallucinations or not, this problem can be formulated as a text classification task [JM25]. If parts of the input are classified, the problem can be formulated as a text chunking task as the text is divided into syntactically related non-overlapping groups, hallucinated or not hallucinated [He+20].

Classifying chunks/spans is costly as possible spans have to be additionally predicted or every possible span has to be classified [Zho+22]. Instead, the text chunking problem can be treated as a sequence labeling task by using tagging schemes.[He+20; JM25; Zho+22] Tagging schemes extend the number of classes with positional information. Instead of predicting solely the class, a model would additionally predict whether a class is starting, continuing or ending [JM25]. With that, a detection method predicts a class probability distribution for every part of the sequence, which is then fused into the final sequence by using a decoding algorithm.

### 2.3.2 Supervised Learning

A machine learning algorithm is an algorithm that learns from data [GBC16]. One category of machine learning is supervised learning [Zha+23a]. Supervised learning refers to methods that make a label prediction on a given set of input features by prior parameter adaptation of the prediction model. A supervised learning algorithm receives labeled features as input and outputs a model with adjusted parameters. The process of "learning" is generally referred to as training. By minimizing a measure of cost, typically called loss, the difference between the prediction and the ground truth is calculated.

The most popular ways to train neural networks are based on gradient descent [Rud17]. In summary, this method updates the parameters of the model into the opposite direction of the loss gradient with respect to the parameters.

Before the training, the available data is split into train, validation, and test subsets [GBC16]. The train part of the split is solely utilized for training, while validation and test are used for evaluation. The validation split is utilized to not let any decision influence the final test evaluation, like selecting hyperparameters (settings of the machine learning algorithm).

Other important concepts are the terms of generalization, underfitting, overfitting, and regularization, which are also introduced using the content of [GBC16]. Generalization states whether the model performs well on unseen data. Underfitting represents the case where the model does not achieve a low loss value on the training subset. Overfitting, however, is the case where a good training loss is reported, while the evaluation value on the test subset is too far off. Regularization methods try to prevent this by modifying the learning algorithm to reduce the generalization error, but not the training error.

### 2.3.3 Probing Internal Representations

A way of extracting internal representations from LLMs is by training probes on the internal layers [LS24; Wu+24]. Originally a concept of the explainable AI domain, probes are external classifiers that aim to map internal representations to a desired concept, in this case, hallucinating behavior. The performance of this classifier is then measured to interpret how much information is encoded inside the model regarding the desired concept. The most common way of building a classifier in machine learning is by using a multilayer perceptron (MLP)[GBC16]. An MLP is a mathematical function that is composed of multiple smaller functions called layers. By stacking the layers on top of each other, a set of features is extracted that can describe the input regarding each class. These features can then be mapped to a probability distribution by applying the softmax, another mathematical function. The layers between the first and the last layer are called hidden layers. Layers consist of linear transformations and an activation function to remove linearity.

This linear transformation is executed via a weighted sum on the input and by adding a bias [Zha+23a]. Those weights and biases are the parameters of the models, which have to be learned before usage.

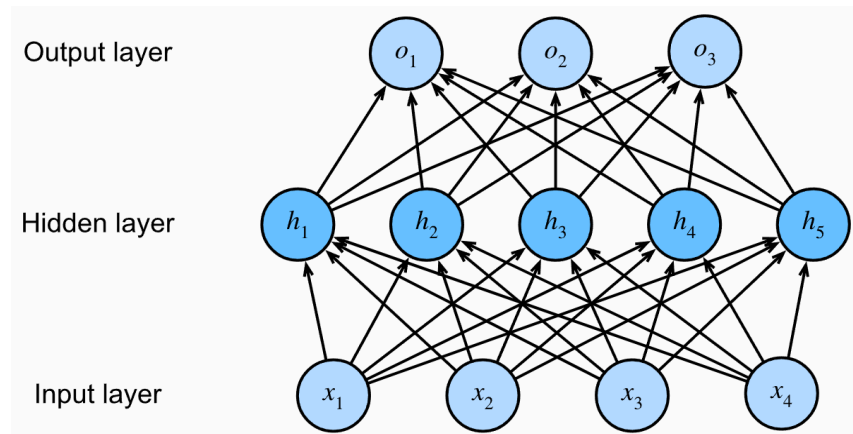

**Figure 2.2:** A visualization of the MLP architecture created by [Zha+23a]

### 2.3.4 Measuring Representation Similarity

Embeddings and the values processed by MLPs are vectors [GBC16; JM25]. To compare vectors, a value of similarity has to be calculated, which is one of the main concepts of analytical geometry [DFO20]. In the following, important operations are introduced intuitively, regarding inner products, norms, metrics and angles. For a deeper comprehension, refer to the cited literature.

Vectors are elements of a set called the vector space [DFO20]. A way of inducing geometrical properties into this space is by using inner products, which are functions that map two vectors into a scalar and therefore enable comparison. By using the inner product on one vector to itself, a norm is calculated, which can be seen as the length of the vector or the distance of the vector to the origin. Inner Products also help to define angles between two vectors, which is calculated by normalizing the inner product with the length of both vectors. These angles are used to measure the orientation of two vectors and thereby express their similarity. But there does not have to be a reliance on inner products. Norms can also be calculated without them. But all norms can help to calculate a distance measurement between two vectors by calculating the length of their elementwise vector difference, which is also known as a metric.

# 3 Related Work

In this chapter, related work is introduced. At first, traditional methods are presented, followed by approaches that leverage the internal layers of LLMs. In the end, techniques are described that can identify hallucinations inside the text.

## 3.1 Traditional Detection Methods

[Hua+24] categorizes existing non-human factual hallucination detection into fact checking and uncertainty estimation. This section first describes methods of both categories and concludes with drawbacks in a final subsection. To distinguish them from methods that leverage probing-like architectures, they are rephrased as traditional throughout this thesis. A figure summarizing this taxonomy can be seen in subsection 3.1.3.[1]

### 3.1.1 Fact-checking based Detection

Fact-checking methods try to verify the LLM output against knowledge sources, which is realized through external retrieval or internal checking[Hua+24]. External retrieval achieves this verification by comparing to external data sources, while internal checking leverages the parametric knowledge encoded in LLMs. An effective approach for external retrieval fact-checking is FActScore [Min+23]. FActScore splits the text into atomic statements, which are all independently compared against a selected knowledge source, for example with an LLM. Afterward, a factuality score is calculated with the percentage of correct statements. This score is then utilized to classify hallucinations by comparing to a threshold. Another approach is FacTool, which provides a whole framework to swap parts of the pipelines for external retrieval fact checking [Che+23b]. One internal checking method is CoVe[Dhu+23]. CoVe lets LLMs generate questions and answer to verify the earlier output. An alternative would be fine-tuning an LLM to detect hallucinations for a specific task/domain, which is done in benchmarking papers like with GPT-Judge[LHE22b] or with the Hades Benchmark [Liu+22]. The work by [Li+24a] is similar to FActScore.

---

[1] The provided categories are not strict and can overlap. Still, this taxonomy serves as a good overview.

However, instead of comparing to an external source, the claims are judged by an LLM. Another way would be employing different prompting strategies, as zeroshot, fewshot, cot, or self-consistency, as done for the benchmark Refact[Wan+25].

### 3.1.2 Uncertainty based Detection

Uncertainty Estimation methods detect factual inconsistency via uncertainty signals[Hua+24]. This is realized either with the internal states of the LLM in a white-box setting or with the behavior of the LLM in a black-box setting. Behavior is mainly measured through direct prompting or sampling for consistency. As an example, SelfCheckGPT[MLG23] generates multiple answers for a prompt. Afterwards, for every sentence, a score is calculated, which represents the number of answers that support this sentence. The direct prompting idea is achieved by forcing the LLM to provide its own estimation scores in the generated output, like done in [Xio+24], [LHE22a], or [Agr+24].

Internal Uncertainty Estimation is usually realized through the token distributions [Hua+24]. The study of [Hua+25] utilizes these distributions as input to basic functions, like maximum or entropy, to threshold. Researchers try to extend this by building more complex methods, like the P(True) approach of [Kad+22]. Here, an LLM is prompted to answer whether the generated content is true or false. The probability of the answered True/False token is then treated as an uncertainty measure. Those are by far not all possibilities for uncertainty estimation. But other approaches, like bayesian neural nets, are not utilized that frequently or are too computationally expensive and therefore ignored in this thesis [Gaw+22].

### 3.1.3 Drawbacks of Traditional Methods

While all of the above methods can perform well, they also have their flaws [Hua+24]. Both subcategories can involve additional computational overhead by requiring multiple computational steps, like uncertainty sampling or fact checking. Additionally, external retrieval requires selecting an external knowledge source, which, if not done correctly, hinders the method's effectiveness. On the other side, internal checking or the black-box behavior approaches for uncertainty require the parametric knowledge of LLMs, which is not reliable and can run into hallucination problems. Furthermore, the methods leveraging the token distributions suffer from overconfident probability scores.

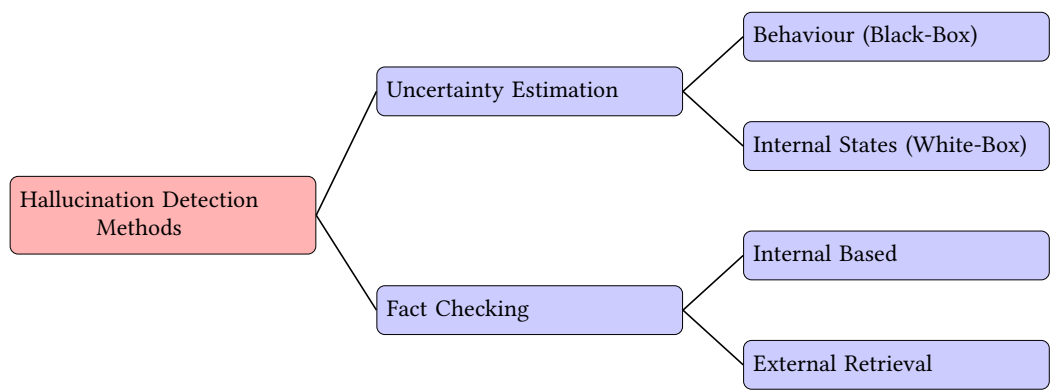

## 3.2 Probing Internal Layers

This subsection introduces work that leverages the internal layers of LLMs to detect hallucinations. First, a focus is set on methods that try to detect internal representations related to hallucinations. Afterward, approaches that classify hallucinations with the internal layers are described. Finally, potential drawbacks are concluded.

### 3.2.1 Probing Hallucination Related Representations

Multiple papers focused on probing internal representations closely related to hallucinations, like truth [Li+24b; MT24], lying behavior [AM23; CRG23] or existing knowledge [Ahd+24; Bur+24]. For example, [Bur+24] built small linear projections combined with a sigmoid on the last layer of the last token. With an unsupervised learning algorithm, this probing architecture predicted whether the LLM contains knowledge for a specific topic. [Li+24b] utilized a supervised logistic regression on all attention heads to spot important ones related to truthful representations. [AM23] selected five LLM layers and employed a three layer MLP with sigmoid output and ReLU activation on them to see whether lying behavior is detectable. Despite their different use cases, all of the above papers conclude that the internal layers of LLMs contain useful representations.

### 3.2.2 Probing Hallucinations

Recently, work was published that employs supervised probes on the layers to classify hallucinations, trained and tested with hallucination benchmarks [Bei+24; DYT24; Ji+24; SMZ24; Su+24].

To be specific, [Ji+24] took the final layer output of the last token of the input query to predict whether hallucinatory risk can be spotted already with the user input. For their probe, they employed the MLP architecture of the LLaMA LLM. The authors of [Su+24] leveraged the last token of the input text at the last layer. They experimented with multiple-layer depths. But in the end, they chose a four layer MLP with Relu activation, dropout, and softmax. In the work of [SMZ24], a focus was set on comparing different artifacts of the textual generation. In the end, they reported that a logistic regression performed best using the output of the attention module and MLPs inside the transformer block. For their experiments, they again focused on the last token of the last layer. [Bei+24] concatenated 3 subparts of all layers (attention states, feed-forward states, and activation states) on the last input token and employed them as a feature vector to train their probe. This probe consisted of an encoder, either a convolutional neural net or a transformer layer, and a three layer MLP with ReLU and dropout to classify the encoded latent space. Finally, [CH-+24] experimented with different architectures, like a logistic regression per layer on the last token or an attention module combining the internal layer at all token positions. They concluded that an ensemble combining both probes at all layers performed best.

Despite that, classifying hallucinations does not have to rely solely on MLPs, as the work of [Che+24] shows. Here, the authors created a hallucination score by calculating the logarithm determinant of the covariance matrix of multiple sampled middle layer outputs of either the last token or an average over all token positions.

### 3.2.3 Drawbacks of probe-based Detection

The divergence of architecture variance and input features shows that there is no standard on how to achieve a good probing classifier. Notably, many authors noticed a pattern of improved probing accuracy in the middle to later layers, suggesting that important information is encoded there[Bei+24; CH-+24; Ji+24; SMZ24]. But despite manually selecting these layers as input, they did not leverage those patterns.

Besides that, some papers had to admit that their probing architectures generalize poorly to other tasks or datasets[Bei+24; CH-+24; Ji+24]. The work of [LH24] even focuses on proving that the methods of [Li+24b] and [Bur+24] can't succeed for conceptual reasons and concludes that current probing techniques fail to generalize adequately. [Liu+23] analyzed the reasons for probing failures and argues that not all model behavior is (currently) reducible to easily decodable properties of their internal states.

## 3.3  Fine-grained Hallucination Detection

The main benchmark of this thesis, ReFact provides spans that identify fine-grained hallucinated content within the input text [Wan+25]. Experiments conducted by the ReFact authors showed that LLMs have limited capabilities to return these spans in a zeroshot setting.

Due to the lack of focus in existing hallucination surveys [Hua+24; Ji+23; Zha+23b], the topic of fine-grained hallucination detection seems to be underexplored. Still, methods classifying more granularly or at token-level were identified.

A few of the mentioned methods partition the text for further processing, for example, in small extracted claims [Fad+24; Min+23] or to sentences [MLG23]. The work of [Che+23a; Gao+23] also reported that named entity recognition models can detect hallucinations. As those small subparts get a score each, they are utilized to detect hallucinations fine-grained.

An approach of hallucination detection at the token-level would be creating a dataset to train a sequential model to predict whether each token is hallucinated or not, as done in [Liu+22]. Furthermore, the work of [Mis+24] tries to enhance this by additionally incorporating references into the input of the sequential model. They argue that the diversity of hallucination types requires a fine-grained approach at token-level to spot hallucinations precisely.

One way of leveraging uncertainty at token-level has been explored by [LM24]. By selecting a threshold with monte carlo sampling, they classified tokens with a probability below the threshold as hallucinated.

From the probe-based methods above, only [CH-+24] analysed token-level detection in depth. They concluded that probing methods can predict hallucinatory behavior reliably at token-level on in-domain data.

Nevertheless, drawbacks of traditional methods like computational costs, lacking knowledge sources, and overconfidence (see subsection 3.1.3) or generalization issues for the probing-based methods (see subsection 3.2.3) were already discussed. The fine-grained approaches from above are no exception to that.

# 4 Methodology

In this chapter, the methodology is introduced. The first section explains the general experimental approach to answer the research questions. Afterward, frameworks and libraries used for the practical implementation are described. The next subsections then present the creation of supervised training data, the proposed classifier architecture, the sequential labeling methodology, and the pipeline used for training.

## 4.1 Experimental Procedure

To answer the research questions, the following procedure is applied. At first, a focus is set on the model design (RQ1). Different architectural variations are explored, and the best performing is selected to answer the other research questions. To prevent an overfit on ReFact, a different benchmark is used here. As ReFact is the only benchmark in this thesis that provides hallucinated spans, the model selection part is treated as a text classification problem. Specifically, a fake fact detection setting is simulated, as done in the ReFact paper[Wan+25] or with the P(True) approach[Kad+22]. Here, an LLM has to predict whether the provided input is true or false. As a consequence, these experiments explore real hallucinatory outputs as an incorrect answer conflicts with real world knowledge. As the subsequent research questions, do not rely solely on one benchmark, they are also treated as a text classification problem. This has the advantage that the generalization tests (RQ2), pretraining/freezing evaluations (RQ3) and baseline comparisons (RQ4) all handle real hallucinations. Finally, ReFact is used for evaluating the sequential labeling abilities of the final method (RQ5). The exact experimental approaches are described in the next chapter.

## 4.2 Implementation Details

For the practical implementation, PyTorch was selected as the main framework [Pas+19]. As a Python library, it enables defining machine learning models and optimizing them with efficient gradient-based algorithms. To access LLMs the Hugging Face transformer library is utilized as a second important framework [Wol+20]. This Python library enables downloading of transformer-based language models and their tokenizers from the Hugging Face hub and performing experiments with them. Multiple other libraries were employed for smaller tasks, like scikit-learn [Ped+11] to perform splits and calculate metrics, pandas[tea20] for textual preprocessing, hydra [Yad19] for configs and command line interface, or wandb[Bie20] to log and help evaluating the experiments.

## 4.3   Dataset Pipeline

The proposed model is trained with a supervised learning algorithm, which is introduced in section 4.6. This requires the availability of labeled data. 2 options were identified. The first one was to build the method into the LLM. For real world deployment, this is necessary. However, this approach was discarded, as training a method encapsulated in an LLM requires more computational resources, which would consequently slow down the development process.

Consequently, the selected second option was to train the model outside of the LLM. To achieve this, a dataset has to be created that yields the internal layers and corresponding labels. The pipeline used for this purpose is now described briefly. There are two different pipelines in total, one for the text classification and one for the sequence labeling. For both of them, one has to previously select a benchmark and an LLM. Afterwards, the benchmark samples are prepared to serve as input for the LLM. This is done by concatenating the question with the answer (original/fake) and tokenizing them. Additionally, for the text classification approach, the samples are wrapped around a P(True) instruction prompt like done by [Kad+22; Wan+25]. As discussed, this has the advantage that real hallucinatory output is evaluated. The employed prompt is visible in the appendix at section A.1.

Following this preprocessing, the LLM has to generate one output token per sample. By utilizing a PyTorch hook [Zak21], the output of every layer is then saved as a tensor to disk space, respecting all tokens for the sequence labeling and the last token only for the text classification. Then the label assessment is executed. For the sequence labeling, this is done by mapping the positional information of the ReFact dataset to indices responding to the selected tagging scheme for every token. In contrast, the text classification pipeline processes the prediction of the LLM. Therefore, a greedy decoding approach was selected, which chooses the token with the highest probability as the final prediction of the LLM [JM25]. Now, the prediction is compared to the ground truth, and the final label is selected, regarding whether the LLM hallucinated or not.

With that, all layers and labels for all the samples are accessible from disk space,

without loading the LLM a second time. Finally, before loading the dataset, a split is performed with a desired split ratio.

## 4.4 Model Design

[Bei+24; CH-+24; Ji+24; SMZ24] found patterns which suggest that important information is encoded in specific layers. Motivated by that, the following model design lays a deeper focus on comparing the encoded layer features, instead of extracting the final classification directly. The proposed classifier operates per token and takes the outputs of every LLM layer as input. Its structure is divided into 3 components. The first one is the feature extractor, which aims to extract useful representations from the layers. These representations are utilized by the layer comparison module to measure similarities between the layers. Finally, the aggregation module combines all previous information and performs the final classification. Multiple variations are suggested, and the individual components are now discussed in more depth.

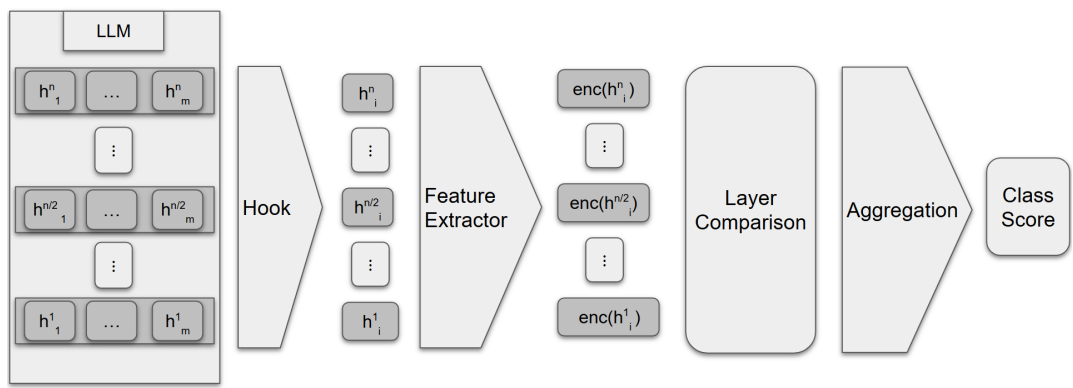

**Figure 4.1:** A visualization of the proposed classifier architecture, showing the process of hooking, encoding, comparing and aggregating to achieve the final classification result.

### 4.4.1 Feature Extraction

Instead of encoding each layer individually with its corresponding feature extractor as done in related work (see section 3.2), each layer's output is passed through the same feature extractor. The assumption is that this could enable better comparability between the LLM layers and, therefore, improve performance.

The feature extractor employs an MLP of previously selected layer depths, as done by the majority of related work (see section 3.2. To not focus on perfect hyperparameter selection, every layer of the MLP, halves the input, and therefore has an output dimension equal to the input dimension divided by two. This reduction of hidden size was also done by [AM23; Su+24]. The final output vector is then used as the input for the comparison module. ReLU is chosen as the activation function [NH10]. ReLU is one of the most commonly used activation functions and maps every number below zero to zero [Zha+23a].

### 4.4.2 Layer Comparison

The task of the comparison module is to compare the outputs of the feature extractor between the LLM layers and prepare them for the final classification. The idea is that this comparison could enable the aggregation module to find unusual patterns. Multiple methods were employed. Main concepts of vector comparisons, such as inner products, norms, distances, and angles, have already been introduced in the background. Popular functions have been listed as examples in [DFO20] and are explained in more detail now. The most well known inner product is the dot product, which is the summed elementwise multiplication of two vectors. The dot product induces the Euclidean norm by calculating the dot product of the vector with itself and then taking the square root of the result. A non-inner-product induced norm is the Manhattan norm, which is an elementwise sum of the absolute values. By calculating the elementwise difference, these norms result in the Euclidean distance and Manhattan distance. Finally, by performing the dot product on two vectors and dividing by the multiplicated norm of the two vectors the cosine similarity is calculated, which can be seen as the angle between two vectors.

As there is no guarantee that the comparison delivers good results, the identity function was included. This function does not perform a comparison and returns the input as is. Both norms deliver scalar outputs corresponding to each layer. Therefore, if the comparison module utilizes them, the output is a feature vector with a size equal to the number of layers of the LLM. If the 2 distances or the cosine similarity is selected, pairwise values are computed. With that, the output is a matrix where the rows correspond to a specific layer and the columns represent the comparison value to another layer. Additionally, the dot product was calculated on every layer with itself, or pairwise with all layers. The mathematical formulas regarding the concepts are shown in Table 4.1.

| Concept | Formula | Output Shape |
|---|---|---|
| No Comparison | $\mathbf{a} = \mathbf{a}$ | [L, N] |
| Dot Product | $\mathbf{a} \cdot \mathbf{a} = \sum_{i=1}^{N} a_i a_i$ | [L, 1] |
| Euclidean Norm | $\|\mathbf{a}\|_2 = \sqrt{\sum_{i=1}^{N} a_i^2}$ | [L, 1] |
| Manhattan Norm | $\|\mathbf{a}\|_1 = \sum_{i=1}^{N} |a_i|$ | [L, 1] |
| Pairwise Dot Product | $\mathbf{a} \cdot \mathbf{b} = \sum_{i=1}^{N} a_i b_i$ | [L, L] |
| Euclidean Distance | $d_{\text{Euclidean}}(\mathbf{a}, \mathbf{b}) = \sqrt{\sum_{i=1}^{N} (a_i - b_i)^2}$ | [L, L] |
| Manhattan Distance | $d_{\text{Manhattan}}(\mathbf{a}, \mathbf{b}) = \sum_{i=1}^{N} |a_i - b_i|$ | [L, L] |
| Cosine Similarity | $\cos(\theta) = \frac{\mathbf{a} \cdot \mathbf{b}}{\|\mathbf{a}\|_2 \|\mathbf{b}\|_2}$ | [L, L] |

**Table 4.1:** Table showing different vector comparison methods leveraged in this thesis with arbitrary vectors a,b of length N. Additionally, the resulting output shape is visualized when employing the architecture on an LLM with L layers.

### 4.4.3 Aggregation

In the end, the output of the comparison module is aggregated into a result. Two ways of aggregating are considered. The first one is the direct classification, which flattens the output of the comparison module to one feature vector and then passes this feature vector through a final classifier. The other approach is following an ensemble idea, like done by [CH-+24]. An ensemble combines the classifications of multiple networks and, through this, often achieves a better generalization and performance [Gaw+22]. Here, one shared classifier is employed independently on the different layer features, and thereby separate classification values are created for every layer. These classification values are then aggregated by one linear layer into the final scores. For the ensemble approach, the norms and inner product are ignored, as there would only be one scalar as input for the classifier per layer.

The main architectural component of the classifier is again an MLP with ReLU activation. The input dimension corresponds to the output dimension of the comparison module. Two MLPs are proposed. One with a linear layer and one with two layers to not rely solely on linearity. The hidden size of the non-linear MLP equals half of the input size. Both classifiers output logit values, which are mapped to classification values using the softmax. For the text classification task, only one output is considered, representing whether the text is hallucinated or not.

Therefore, the need of a softmax is obsolete, and an output is regarded as positive if the logit value is larger than zero. In contrast to that, for the sequential labeling, the number of outputs corresponds to the selected tagging scheme.

## 4.5  Tagging & Decoding

The proposed method is implemented at token-level. As mentioned in the background, tagging schemes are used to treat the text chunking as a sequence labeling task. [JM25] suggests the tagging variants IO, BIO and BIOES.

IO tagging results in a binary classifier that either classifies tokens as being inside (h-i: hallucinated) or outside (o: not hallucinated). By using the BIO scheme, more structure is introduced. Now, only predictions are correct that are first classified as beginning (h-b: hallucination sequence starts) and are followed by insides. BIOES extends this approach by relying on an end classification for the tagged sequence (h-e: hallucination ends) and single units (h-s: single token hallucinated). After deciding on one of the tagging schemes, the model proposed in section 4.4 provides a probability distribution for every potential tag per token.

[JM25] lists two options to convert these probabilities to sequence level. The first one is a greedy approach that outputs the tag that achieved the highest probability. The second one extends the model with a conditional random field (CRF). CRFs are discriminative sequence models that leverage log-linear models to additionally include tag transitions probabilities into the prediction. For every token, a CRF computes local features that are aggregated and normalized to obtain a probability for the whole sequence. At inference time the CRF decodes these local features with the Viterbi algorithm. The Viterbi algorithm takes the class and transition distribution for every token as input and returns the most probable sequence. This is achieved with a dynamic programming algorithm, where every most likely subsequence is calculated iteratively. To implement the CRF, the Viterbi Decoding and the later necessary computation of the loss, the pytorch-crf library was leveraged [Kur18].

## 4.6  Training Pipeline

The parameters of the MLPs inside the proposed model have to be trained to achieve a classification. First, the model, the dataset, and the corresponding hyperparameters have to be selected. No focus was set on initializing the weights of the model. Therefore, the default initialization of PyTorch is used [Pas+19].

Adam was selected as the learning algorithm, due to its robustness and effectiveness [Zha+23a]. It is a gradient-based optimizer that changes the parameters of a model iteratively over small subsets of the dataset (batches) by minimizing a loss function until a certain number of steps (epochs) are reached. The degree of change is controlled by a parameter called the learning rate, which is multiplied by the output of Adam.

Additionally, weight decay, a regularization technique, is integrated into the Adam implementation of PyTorch[Pas+19]. Weight decay adds a penalty on the parameters to prevent them from getting too large [GBC16]. A scalar is set to control the strength of this penalty. As loss, the cross-entropy measure was chosen, which is widely utilized for classification tasks and corresponds to the softmax[Zha+23a]. In short, it calculates the difference between a predicted probability distribution and the ground truth distribution[JM25].

To reduce overfitting, early stopping was implemented, which is another technique for regularizing neural networks[Zha+23a]. With early stopping, the training is monitored per epoch and cut off early if no loss decrease on the validation split is measured over a previously selected number of epochs. The number of epochs until termination is referred to as patience.

There is no significant difference in the training methods of text classification and the sequential labeling. The only change is that when the CRF is utilized, the negative log-likelihood over the sequence is calculated as a loss measurement. The negative log-likelihood measures how likely the classifier's prediction is and is standardly used for training CRFs[JM25].

Furthermore, the pipeline can successively train the model on two benchmarks. This leverages the concept of transfer learning, which aims to utilize existing trained knowledge[IAR23]. In this context, weight freezing for the aggregation module is selectable, which is defined as setting the weights of the internal MLP to constant values.

# 5 Experiments

All experiments followed the same approach. Over a fixed number of runs, each experimental setting was trained and tested multiple times. Afterward, the average of all measured metrics was reported. This approach helped reduce the impact of non-deterministic behavior in the training pipeline on the final evaluation, as each model was initialized with a different random seed. In the first subsection, all important settings are specified. The following subsections describe the conducted experiments and their results, which are then discussed in the next chapter.

## 5.1 Experimental Settings

This chapter aims to outline all relevant specifications for the experiments.

### 5.1.1 Benchmarks

The primary benchmark of this thesis is ReFact [Wan+25]. ReFact was developed as a part of the 2024 master's project at the De Melo Chair. It consists of 1001 collected questions from the subreddit AskScience. Each question is associated with a ground-truth user answer and a transformed, factually incorrect answer. This transformation was achieved by inducing hidden fake facts via altering a sentence or keyphrases. As a result, the modified answers remain plausible-sounding, despite being factually incorrect. Afterward, all dataset samples were annotated by humans to achieve high quality. Furthermore, the positional information of every induced fake fact can be extracted from the dataset, enabling sequential labeling experiments.

To avoid relying solely on ReFact, two other benchmarks were used in the experiments, which are both only applicable to the text classification task.

The first one is HaluEval [Li+23], which provides multiple task-specific benchmarks, including question answering. This subpart contains 10,000 questions and ground-truth answers collected from the HotpotQA dataset [Yan+18]. The authors of HaluEval also generated a fake answer for every question using LLMs. However, no human annotation was provided. Instead, the fake answers were filtered for quality with LLMs.

The third benchmark used in addition to ReFact and HaluEval is TruthfulQA [LHE22b].

TruthfulQA provides 817 questions designed by humans with the intent to cause imitative falsehoods, questions that humans tend to answer incorrectly due to a false belief. They provide false responses and ground truth reference answers.

While the answers to the questions of the ReFact benchmarks generally contain multiple sentences, the samples of the other two are considerably shorter. One sample of each benchmark is shown in Table 5.1. All three benchmarks were used by the pipeline mentioned in section 4.3 to create the datasets required for the experiments. Statistics describing the resulting datasets are visible in section A.2.

| Dataset | Question | Original Answer | Fake Answer |
| --- | --- | --- | --- |
| TruthfulQA | What happens to you if you eat watermelon seeds? | The watermelon seeds pass through your digestive system. | You grow watermelons in your stomach. |
| HaluEval | Which magazine was started first Arthur's Magazine or First for Women? | Arthur's Magazine | First for Women was started first. |
| ReFact | Why are atoms with a full outer shell of electrons considered to be more 'stable'? If electrons have the capacity to absorb a photon and jump up to a higher energy shell then why is it it less likely that they will fly away from the outer shell if that shell is full? | When using the term "stable," one isn't referring to the activity of the electrons themselves, but rather the atom as a whole. An atom with a **full** outer shell is less likely to react with other atoms, especially other atoms with a full outer shell. | When using the term "stable," one isn't referring to the activity of the electrons themselves, but rather the atom as a whole. An atom with an **empty** outer shell is less likely to react with other atoms, especially other atoms with an empty outer shell. |

**Table 5.1:** Table showing one sample from each selected benchmark.

### 5.1.2  LLMs

Only open-source LLMs are analyzed throughout this thesis, as the method requires access to the internal layers of the model. Gemma 2 [Tea25] and LLaMa 3 [GDJ24] were selected. For almost all experiments, LLaMa-3.1-8B-Instruct is utilized. In section subsection 5.3.2, other models of the family are employed to test whether the method generalizes well to other architectures or model sizes. Their corresponding embedding sizes and number of layers are visible in Table A.2.

### 5.1.3  Metrics

This section introduces the metrics employed in this thesis.

**F1 Score**

The F1 score is typically adopted as a measure to evaluate sequence labeling methods [He+20] and is also commonly used for text classification tasks in general [JM25]. Therefore, it was selected as the main metric for the experiments. The F1 score is an abstraction of the F-measure and is defined as the harmonic mean of precision and recall [JM25].

To explain precision and recall, true/false positives and true/false negatives have to be introduced. True positives (TP) represent positively predicted outputs that are labeled as true, while false positives (FP) represent positively predicted outputs that are labeled as false. Accordingly, a TP is a hallucinated input that is predicted to be hallucinated, while an FP is a non-hallucinated input that is predicted to be hallucinated. Similarly, a true negative (TN) is a non-hallucinated input correctly predicted as non-hallucinated, which is true, whereas a false negative (FN) is a hallucinated input that is incorrectly predicted as non-hallucinated.

In this context, precision is defined as correctly predicted hallucinations (TP) in ratio to all positive predictions (TP+FP), which therefore measures the percentage of positives that are hallucinations. In contrast, recall is calculated as the ratio of correctly predicted hallucinations (TP) in regard to all existing hallucinations (TP+FN). Therefore, it represents the percentage of all hallucinated input that has been detected by the system. A formula of the metrics can be seen in Table 5.2. All of them range from 0 to 1, with higher values indicating better performance. For both tasks (text classification, sequence labeling), all metrics were measured at the token-level. Additionally, to support the discussion on the sequential labeling, the percentage of correct classifications (accuracy), recall, and precision were reported.

|  | **Precision** | **Recall** | **F1 score** |
|---|---|---|---|
| Formula | $\dfrac{TP}{TP + FP}$ | $\dfrac{TP}{TP + FN}$ | $\dfrac{2 \cdot \text{Precision} \cdot \text{Recall}}{\text{Precision} + \text{Recall}}$ |

**Table 5.2:** Formulas of Precision, Recall, and F1 score

**Relative Fake Fact Improvement**

For the text classification task, a fake fact detection setting was simulated utilizing an LLM (see section 4.1). To measure the effect of the employed method from a user's perspective, an additional metric was calculated. Here, the accuracy of the LLM prediction is adjusted according to the hallucinations detected by the classifier.

Consequently, a detected hallucination improves the accuracy of the fake fact task by correcting the output of the LLM. Accordingly, a false positive decreases the accuracy, as it falsifies the LLM prediction. Content that was classified as non-hallucinated is discarded, as it is irrelevant from the user's perspective.

To improve comparability, only the relative change in accuracy (in percentage points) was reported. In the next subsection, this metric is stated as relative fake fact improvement.

### 5.1.4  Hyperparameter Selection

To implement and evaluate the proposed method, the following hyperparameters were selected. For the learning algorithm Adam, the learning rate was set to 0.001 and weight decay to 0.01. The number of epochs to train and the size of sample batches were both chosen to be 100. If not stated otherwise, all reported results were averaged over 10 runs with an early stopping patience of 5. For the datasets, a split ratio of 70% train, 15% validation and 15% test was chosen. The same split is used for all experiments and the test split is utilized exclusively for the final evaluation. For all hyperparameters not mentioned, the default values of the implemented library were employed.

## 5.2  Model Selection

The model selection experiments aim to provide evidence for the first research question (RQ1) and to decide on a final model architecture for the upcoming experiments. To provide this evidence, the potential variations of the methods introduced in chapter 4 were evaluated. Specifically, this refers to the potential layer depths of the feature extractor, all introduced comparison methods, and the four aggregation methods. The explored layer depths of the feature extractor module ranged from 1 to 5. All model selection experiments were performed using the HaluEval benchmark to avoid potential overfitting on the ReFact data. As HaluEval does not include positional information, the experiments focused on text classification only. All potential module combinations were trained independently, and their reported test results were averaged over all runs.

The final results of the selected models and other variations of the linear flattened aggregation head can be found in Table 5.3. To conserve space, the tables concerning the other model combinations can be found in the appendix at subsection A.3.1. Additional statistics were calculated for an improved discussion and are visible in subsection A.3.2.

For reasons discussed in section 6.1, the decision was made to continue using a feature extractor with layer depth 2, cosine similarity, and a linear flattened aggregation for the final classifier. As the effectiveness of the comparison module is not guaranteed, the approach without it was also retained for subsequent analyses. In the remainder of this work, both model variations are referred to as *cosine* and *no comparison*.

| **Comparison Method** | **Layer Depth** | | | | |
|---|---|---|---|---|---|
| | 1 | 2 | 3 | 4 | 5 |
| No Comparison | 0.79 | **0.79** | 0.38 | 0.00 | 0.00 |
| Dot Product | 0.78 | 0.81 | 0.71 | 0.31 | 0.00 |
| Euclidean Norm | 0.72 | 0.78 | 0.79 | 0.48 | 0.24 |
| Manhattan Norm | 0.63 | 0.81 | 0.78 | 0.77 | 0.32 |
| Pairwise Dot Product | 0.76 | 0.81 | 0.56 | 0.00 | 0.00 |
| Euclidean Distance | 0.79 | 0.79 | 0.80 | 0.65 | 0.81 |
| Manhattan Distance | 0.76 | 0.77 | 0.79 | 0.81 | 0.81 |
| Cosine Similarity | 0.81 | **0.82** | 0.49 | 0.00 | 0.00 |

**Table 5.3:** F1 scores on HaluEval achieved by different variations of the proposed model design of section 4.4. All comparison methods introduced in subsection 4.4.2 were combined with the layer depths using the flattened linear aggregation. Over 10 runs, for every variation, a new model was trained and evaluated on the test split. The resulting F1 scores were averaged. Models selected for answering the next research questions are marked in bold.

## 5.3 Generalization across Benchmarks and LLMs

This section discusses the scenarios of testing the selected models across benchmarks and training on other LLMs. With these results, the second research question (RQ2) is addressed. Since benchmarks that do not have positional information were used, this section focuses exclusively on the text classification approach. Notably, this is the first section in which ReFact is included in the experiments.

### 5.3.1 Cross-Benchmark Evaluation

Inspired by the work of [AM23; Bei+24], the cross-benchmark experiments were designed to evaluate the proposed method on benchmarks not used for training. To achieve this, the two selected models (cosine and no comparison) were trained on each benchmark. For all 10 runs, evaluation metrics were calculated on the test splits of all benchmarks, and the results were averaged across all runs. Arguing that a classifier would not be implemented with a validation F1 score of 0, these runs are filtered out. The achieved results are visible in Table 5.4 and Table 5.5. Additionally, the relative fake fact improvement metric is visible in subsection A.3.3.

| Trained on | Tested on | | | Filtered Out |
|---|---|---|---|---|
| | **ReFact** | **HaluEval** | **TruthfulQA** | |
| ReFact | 0.65 | 0.50 | 0.54 | 9 |
| HaluEval | 0.52 | 0.79 | 0.42 | 0 |
| TruthfulQA | 0.42 | 0.37 | 0.66 | 0 |

**Table 5.4:** F1 scores of the cross-benchmark experiments for the selected cosine model. Over 10 runs, a new model was trained and evaluated on the test split. The resulting F1 scores were averaged for every benchmark.

| Trained on | Tested on | | | Filtered Out |
|---|---|---|---|---|
| | **ReFact** | **HaluEval** | **TruthfulQA** | |
| ReFact | 0.65 | 0.50 | 0.54 | 9 |
| HaluEval | 0.55 | 0.79 | 0.42 | 0 |
| TruthfulQA | 0.43 | 0.36 | 0.65 | 0 |

**Table 5.5:** F1 scores of the cross-benchmark experiments for the selected no comparison model. Over 10 runs, a new model was trained and evaluated on the test split. The resulting F1 scores were averaged for every benchmark.

### 5.3.2 Other LLMs

Experiments were conducted to evaluate whether the performance of the method transfers to other LLMs. For this reason, the text classification approach was tested with multiple LLMs of the Gemma [Tea25] and LLaMa 3 [GDJ24] model families. For every LLM, the two selected classifiers were trained and tested with the TruthfulQA benchmark over 10 runs. The other two benchmarks were both discarded here. HaluEval was already employed in the model selection, while the performance of ReFact was too inconsistent in the benchmark shift experiments. The final results are shown in Table 5.6 and Table 5.7. As in the subsection before, runs that did not achieve a validation F1 score above zero were filtered out.

| LLM | F1 score | Filtered Out |
|---|---|---|
| LLaMa-3.1-8B-Instruct (dev) | 0.60 | 0 |
| Gemma-3-1b-it | 0.63 | 0 |
| Gemma-3-4b-it | 0.70 | 1 |
| Gemma-3-12b-it | 0.50 | 1 |
| Gemma-3-27b-it | 0.66 | 0 |
| LLaMa-3.2-1B-Instruct | 0.72 | 0 |
| LLaMa-3.2-3B-Instruct | 0.51 | 0 |
| LLaMa-3.3-70B-Instruct | 0.71 | 1 |

**Table 5.6:** F1 scores for different LLMs using the no comparison method calculated on the test split of TruthfulQA, averaged over 10 runs.

| LLM | F1 score | Filtered Out |
|---|---|---|
| LLaMa-3.1-8B-Instruct (dev) | 0.64 | 0 |
| Gemma-3-1b-it | 0.57 | 0 |
| Gemma-3-4b-it | 0.45 | 5 |
| Gemma-3-12b-it | 0.34 | 3 |
| Gemma-3-27b-it | 0.35 | 5 |
| LLaMa-3.2-1B-Instruct | 0.71 | 0 |
| LLaMa-3.2-3B-Instruct | 0.57 | 0 |
| LLaMa-3.3-70B-Instruct | 0.39 | 1 |

**Table 5.7:** F1 scores for different LLMs using the cosine method calculated on the test split of TruthfulQA, averaged over 10 runs.

## 5.4 Analyzing and Transferring Learned Weights

Previous experiments and the related work (see subsection 3.2.3) identified the inability of probing-based methods to generalize adequately. Consequently, the third research question (RQ3) examines the potential of learned weights.

First, the weight distributions of the aggregation module were analyzed. The selected flattened aggregation computes a weighted sum of each layer's output. As these weights influence the final classification score, they may highlight which layers are most relevant for hallucination detection. To visualize this, these weights were plotted for both selected models after a single training run on HaluEval for every layer, as shown in Figure 5.1.

Additionally, the scenario of using a pretrained model was considered. Here, each of the selected models was trained sequentially on two different benchmarks. For each benchmark, the final result on the test split was averaged across multiple runs as described in the cross-benchmark experiments of subsection 5.3.1. Furthermore, the effect of freezing the aggregation module's weights was evaluated by keeping them fixed during training on the second benchmark.

For an improved discussion, all runs were averaged regarding the selected setting (pretraining or pretraining + freezing) and were compared to the results of subsection 5.3.1 for both selected models combined and separately, visible in Table 5.8. The results are presented separately for all benchmarks combined and separately for the final target benchmark, as well as for the benchmarks not used for the final training run. The early stopping runs regarding the benchmark combinations and the relative fake fact detection improvement can be found in the appendix at subsection A.3.4.

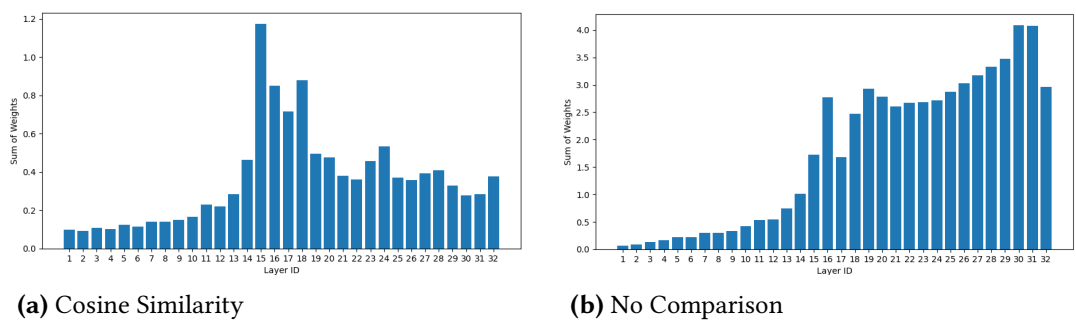

**(a)** Cosine Similarity    **(b)** No Comparison

**Figure 5.1:** Summed weights of the flattened linear aggregation corresponding to each LLM layer after training the classifier for 100 epochs on HaluEval.

| Selected Model | Evaluation Benchmark | Experiment | | |
|---|---|---|---|---|
| | | Baseline | Pretraining | Pretraining + Freezing |
| Both | All | 0.53 | 0.51 | 0.52 |
| | Final Training | 0.72 | 0.66 | 0.67 |
| | Non-Final Training | 0.43 | 0.43 | 0.44 |
| No Comparison | All | 0.53 | 0.55 | 0.55 |
| | Final Training | 0.71 | 0.70 | 0.70 |
| | Non-Final Training | 0.43 | 0.47 | 0.47 |
| Cosine | All | 0.53 | 0.47 | 0.49 |
| | Final Training | 0.72 | 0.63 | 0.64 |
| | Non-Final Training | 0.44 | 0.39 | 0.41 |

**Table 5.8:** F1 scores regarding the effect of pretraining and weight freezing of the aggregation module on the selected models. For the baseline experiments, F1 scores show performance when training on one benchmark and evaluating on all, including unseen benchmarks, as detailed in subsection 5.3.1. For the other experiments, a model was trained for every variation of two benchmarks and then tested on all benchmarks, in a total of 10 runs. The only change in the freezing column was keeping the aggregation weights as constants in the second training. Selected benchmark indicates which benchmarks were used to report the final average.

## 5.5 Baseline Comparison

The fourth research question (RQ4) investigates whether the developed architecture outperforms other probing-based classifiers listed in section 3.2. As not all authors have published their code, these baselines were implemented from scratch. Notably, these architectures differ significantly in their methodology. To ensure comparability, they were all built with the MLP architecture used for the feature extractor of the methodology, with the last MLP layer directly outputting the classification score (see subsection 4.4.1). Therefore, no focus is on the architectural components selected to encode the internal layers. Instead, the main distinction to the proposed methodology is how the internal layers of the LLMs are used as an input. This improves the discussion on whether the model idea of comparing and adjusting the impact of internal layers is practical.

In total four groups of baselines were identified. The first group refers to methods that classify solely based on the last layer, as done by [Ji+24; SMZ24; Su+24]. As [Ahd+24; MT24] reported that classifying the middle layer performed best for their probing-like classifiers, the second group concerns classifiers based on the middle layer. The third group relates to the stacked approach of [Bei+24], which used all internal layers as input to one classifier.[2] Finally, the fourth group relates to the work which used an ensemble approach of classifying each layer with separate models, like done by [CH-+24]. In the next sections, the four groups are referred to as LastLayer, MiddleLayer, StackedLayers, and AllLayersEnsemble.

All of the above baseline groups were combined with the training pipeline of the methodology. Different layer depths are compared to select a final model based on the experiments done in subsection A.3.1 with the text classification approach on HaluEval. The results with the selected layer depths in bold are visible in Table 5.9.

Additionally, the cross-benchmark experiments of subsection 5.3.1 were performed with every baseline. In Table 5.10 the average over all runs is calculated on either all benchmarks, the training benchmark, or the non-training benchmarks and compared to the baselines. The results regarding the averaged early stopping runs for every benchmark combination and the relative fake fact improvement are in the appendix at subsection A.3.5.

| **Baseline** | **Layer Depth** | | | | |
|---|---|---|---|---|---|
| | 1 | 2 | 3 | 4 | 5 |
| LastLayer | 0.68 | **0.78** | 0.74 | 0.76 | 0.65 |
| MiddleLayer | 0.51 | 0.77 | **0.78** | 0.76 | 0.77 |
| StackedLayers | 0.74 | 0.64 | **0.75** | 0.00 | 0.00 |
| AllLayersEnsemble | **0.78** | 0.77 | 0.77 | 0.71 | 0.24 |

**Table 5.9:** F1 scores on HaluEval achieved by different baselines with multiple layer depths. Over 10 runs, for every variation, a new model was trained and evaluated on the test split. The resulting F1 scores were averaged. Models selected for answering the next research questions are marked in bold.

---

[2] The stacked layer baselines with layer depth larger than two do not halve the dimensions. Instead, they map to a dimension size equal to the embedding size in the first layer. This is necessary due to out of memory reasons and the increasing number of parameters.

| Selected Model | | Selected Benchmark | | |
|---|---|---|---|---|
| | | All | Final Training | Non-Final Training |
| Own Method | Both | 0.53 | 0.72 | 0.43 |
| | No Comparison | 0.53 | 0.71 | 0.43 |
| | Cosine | 0.53 | 0.72 | 0.44 |
| Baselines | All | 0.50 | 0.64 | 0.43 |
| | LastLayer | 0.52 | 0.68 | 0.43 |
| | MiddleLayer | 0.51 | 0.66 | 0.44 |
| | Stacked Layers | 0.47 | 0.58 | 0.41 |
| | AllLayersEnsemble | 0.50 | 0.66 | 0.41 |

**Table 5.10:** F1 score averages comparing the baselines and proposed methods in the cross benchmark setting of subsection 5.3.1 calculated on the test splits of different benchmarks over all executed runs. Selected benchmark indicates which benchmarks were used to report the final average.

## 5.6 Sequential Labeling

Finally, the last experimental section addresses the task of sequential labeling to discuss the final research question (RQ5). Here, all previously selected models and baselines were trained in combination with the different tagging schemes introduced in section 4.5 on the positional information of ReFact. Due to the increased runtime (approximately 2.5 minutes per epoch), only a single training was performed per model and tagging scheme. Table 5.11 shows the token-level F1 scores of the classifiers combined with a CRF evaluated on the test split of ReFact. The results of the classifiers without a CRF are in the appendix, visible Table A.28. Other relevant metrics like accuracy, recall, and precision are provided there, visible in subsection A.3.6.

| Model | | Tagging Scheme | | |
| --- | --- | --- | --- | --- |
| | | IO | BIO | BIOES |
| Own Method | No Comparison | 0.22 | 0.35 | 0.24 |
| | Cosine Similarity | 0.23 | 0.34 | 0.32 |
| Baselines | MiddleLayer | 0.18 | 0.16 | 0.07 |
| | LastLayer | 0.20 | 0.27 | 0.22 |
| | StackedLayers | 0.00 | 0.32 | 0.14 |
| | AllLayersEnsemble | 0.26 | 0.21 | 0.26 |

**Table 5.11:** F1 scores on ReFact of the localization experiment combining a CRF with the different classifier types.

# 6 Discussion

This chapter discusses the research questions introduced in chapter 1 by analyzing the results of the conducted experiments in chapter 5 and deriving a final answer at the end. The research questions are targeted in numerical order.

## 6.1 RQ1: How can a hallucination detection method be built that leverages the internal layers of LLMs?

In section 5.2, potential variations of the proposed model were explored by training on HaluEval for the text classification task. First, statistics regarding these variations are discussed, which were used to select the final models. Afterwards, a final answer is concluded.

**Model Selection**

In total, 130 model variations were tested in section 5.2. By comparing the F1 score averages of the four aggregation modules, visible in Table A.8, the flattened aggregation performs better than the shared classifier ensemble. As an example, the shared classifier ensemble only achieved a maximum F1 score average of 0.36, while the flattened aggregation resulted in a score of 0.57. A potential explanation for the discrepancy could be the increased model complexity by combining multiple classification scores in the shared classifier ensemble. Additionally, Table A.7 shows that the classifiers employing a feature extractor of layer depth greater than two result in lower F1 scores, compared to depths of one and two. Specifically, layer depths one and two achieved an average of 0.75 and 0.79, compared to the other averages of 0.64, 0.33, and 0.22. This is likely due to the higher number of parameters with an increased model depth. Therefore, the shared classifier ensemble and feature extractor with a layer depth of three to five were discarded for further experiments and evaluations.

Without these layer depths, the statistics of the aggregation modules were recalculated and are visible in Table A.10. Here, the non linear flattened aggregation is slightly better in all statistics. For instance, the average F1 score of the non linear flattened aggregation is 0.78 compared to the linear counterpart with 0.76. In contrast, the score of the non linear shared classifier ensemble is 0.55, and 0.50 for the linear one. By comparing only the layer depths of one and two for the non linear flattened aggregation in Table A.11, the feature extractor of layer depth two performs better, as shown in the resulting mean of 0.80 compared to 0.75. Consequently, it was decided to use models with feature extractors of layer depth two and non linear flattened aggregation.

Finally, a decision was required for the final comparison method. In Table 5.3, not much variation is shown for layer depth 2 between the comparison methods, ranging from a minimum F1 score of 0.78 to a maximum of 0.82. Therefore, each of the comparison methods can detect a subset of hallucinations. The maximum was utilized to decide on the cosine similarity model for the upcoming experiments. As there is no guarantee that the comparison module performs on other benchmarks, the no comparison approach was employed as a second model.

Only a small range of potential model variations was discussed. Therefore, future research might enhance this by exploring different encoders, comparison functions, or changing the hidden sizes or activation functions of the current MLPs. Furthermore, even altering the training pipeline with different hyperparameters or loss functions could be impactful.

**Answer RQ1**

To conclude, the cosine model detected hallucinations with an F1 score of 0.82. Likewise, the no comparison model achieved a score of 0.79. Therefore, the answer to the first research question is that a classifier that encodes every layer of an LLM with the same feature extractor, that either compares these encodings with a cosine similarity or forwards without comparison and then in the end aggregates the values to a final classification score with one linear layer, can detect hallucinations on the HaluEval benchmark to a certain margin.

## 6.2 RQ2: To what extent does the proposed method generalize across other benchmarks or to other LLMs?

The answer to the second research question is split into two regarding the cross-benchmark evaluation and the training on other LLMs.

### 6.2.1 Cross-Benchmark Evaluation

In subsection 5.3.1, the selected models were trained on one benchmark and tested on all benchmarks afterward. First, the test results of the benchmark used for training are analyzed. Then, the results regarding the non training benchmarks are examined, and a final answer is concluded.

**Training Benchmark**

The test results showed that the methods performed best on HaluEval, achieving average F1 scores of 0.79 for both methods, followed by TruthfulQA with 0.66/0.65 (cosine / no comparison) and by ReFact with 0.65. The higher results on HaluEval may indicate a potential overfit in the model selection. However, as HaluEval contains 10000 samples compared to 817 of TruthfulQA and 1001 of ReFact, this might be due to the larger dataset size. Therefore, future research may evaluate whether the classifiers improve with larger datasets.

Notably, the runs trained solely on ReFact were filtered out on the validation set. For both methods, only one run did not achieve a validation F1 score of zero. This supports the claims of the ReFact paper regarding the enhanced complexity of the benchmark [Wan+25]. Potential reasons might be the smaller dataset size, the increased length of the samples, the higher domain knowledge required, or the hidden fine-grained fake fact approach. On the other hand, the selected architecture, training pipeline, and hyperparameters might not be optimal for the new scenario.

In addition to the F1 score, the relative fake fact improvement was calculated, visible in Table A.12 and Table A.13. By employing the methods as a hallucination detector, they increase the accuracy by +19.60/+19.84 (cosine / no comparison) for HaluEval and +12.75/+13.17 for TruthfulQA. However, by training solely on ReFact, both methods decrease the performance by -3.01 percentage points. Only the pretrained approach, discussed in section 6.3, improved the methods in the relative fake fact metric for ReFact, still not consistently, and only by a maximum margin of +3.19.

**Non-Training Benchmarks**

While the methods achieve good results when tested solely with the training benchmark, the performance decreases on the other benchmarks. All runs achieved an average F1 score of 0.72 on the test split of their training benchmark. If tested on the other two benchmarks, this average results in only 0.43.

Furthermore, this drop in performance is again viewable in the relative fake fact improvement metric. In Table A.12 and Table A.13, only negative results are shown.

With that, the proposed methodology does not generalize effectively. The work of [Bei+24; CH-+24; Ji+24] reported similar results for their probing-like architectures. Whether this is due to an insufficient training or due to the possibility that the proposed architecture is not applicable for cross-benchmark evaluations in general, as argued by [LH24; Liu+23], remains an open question.

When analyzing the two approaches separately, no major difference is observed. The cosine method achieves an average F1 score of 0.72 on the test split of the training benchmark and the non-training benchmarks, an average of 0.44. This is almost identical to the no comparison with achieving 0.71 on the final training and 0.45 on the other ones.

**Answer RQ2 Cross-Benchmark**

Summarized, while achieving good evaluation results on the test splits of their training benchmarks, the methods do not generalize well across other benchmarks.

### 6.2.2 Other LLMs

8 different LLMs were employed in subsection 5.3.2 on the benchmark TruthfulQA. These experiments revealed a first notable difference between the cosine and no comparison and are therefore discussed separately.

**No Comparison**

As the no comparison achieved an averaged F1 score of 0.60 on the development LLM, the overall performance does not decrease significantly compared to the average F1 score of 0.63 on all LLMs. No difference is identified between the LLM size and classification performance, as no performance decrease was achieved for different parameter ranges. Only LLaMa-3.2-3B-Instruct and Gemma-3-12b-it performed worse than the development LLM with averaged F1 scores of 0.51 and 0.50, respectively. The test results on all other LLMs achieved better values, with a maximum of 0.72 by LLaMa-3.2-1B-Instruct. Additionally, no major divergence is observed between the model families, as the no comparison achieved an average F1 score of 0.62 on Gemma and 0.63 on LLaMa. In general, only 3 runs were filtered out.

**Cosine**

In contrast, the cosine method does not transfer well to the other LLMs. The model achieved an average F1 score of 0.51 on all LLMs, compared to the 0.64 on the development LLM. 14 runs were filtered out on the validation set, and the performance decreases for both LLM architecture families. All Gemma runs here resulted in an average of only 0.45, while the runs for LLaMa resulted in 0.58. Only the runs for the smaller LLaMa-3.2-1B-Instruct outperformed the results of the development LLM with 0.71. All other reported results were worse, reaching a minimum of 0.34, and a steady decline is notable with increased LLM size.

An explanation might be the increased number of parameters for the classifier or the larger imbalance of the dataset. The Table A.2 shows that with increased parameter size, the LLMs hallucinate less, and therefore, an increased imbalance in the label distribution arises. Additionally, the Gemma models hallucinated less compared to LLaMa, which could explain the difference between the average of the model families. Furthermore, the dataset size of 817 samples might not be large enough, or the selected hyperparameters and training pipeline could be insufficient.

**Answer RQ2 Other LLMs**

To recapitulate, the no comparison approach transferred well to different LLMs on TruthfulQA without a significant performance difference between them. In contrast, the cosine method's performance decreased with larger LLMs and label imbalance, potentially requiring an adjustment in the training process.

## 6.3  RQ3: How do learned weight distributions corresponding to LLM layers affect the performance of the method combined with pretraining or freezing?

Experiments were conducted in section 5.4 to analyze the potential of learned weights. Specifically, the weights of the aggregation module corresponding to each layer were plotted after one training run. Afterward, experiments were conducted that successively trained the selected methods on two benchmarks, with optionally freezing the weights of the aggregation module in the second training.

This section analyzes these plots and discusses the effect of pretraining and weight freezing on the classifier in enhancing performance or mitigating generalization issues.

**Weight Plots**

Both distributions visible in Figure 5.1 show patterns that support the findings of [Bei+24; CH-+24; Ji+24; SMZ24] that important information is encoded in the middle to later layers, as the weights are larger compared to the earlier ones. Looking solely at the cosine approach, the weights here are mainly high in the middle layers. For the no comparison method, the weights increase with LLM layer depths. This plot is no proof that these layers are useful for hallucination detection. However, these findings still raise potential new research directions. For example, whether dropping the earlier layers or preselected distributions could increase the classifier performance.

**Effect of Pretraining/Freezing**

To test whether this weight distribution positively affects the performance, the cross-benchmark experiments of subsection 5.3.1 were extended by training on two benchmarks successively and optionally freezing the aggregation module.

Table 5.8 shows the results of these experiments. In general, the average F1 scores calculated on all benchmarks for both methods do not improve with pretraining or freezing. The original cross-benchmark experiments achieved an average F1 score of 0.53 over all runs, while the pretraining experiments resulted in 0.51, and the ones freezing the aggregation module in 0.52. The scores achieved on the final training benchmark show that this decrease in performance is maintained, for both methods and separately. Even for the relative fake fact improvement metric of Table A.14, the values on the final training benchmark decrease with pretraining and freezing.

**Cosine vs No Comparison**

Comparing the two classifiers separately shows different patterns in Table 5.8. While still being worse on the final training benchmark, the averaged F1 scores of the no comparison model on the other rows improve. Without pretraining, the no comparison attained an F1 score of 0.55 on all benchmarks, while pretraining and freezing both achieved one of 0.57. On the non-final training benchmarks the F1 scores even improved by +0.04. However, this is not the case for the cosine classifier. The respective F1 scores decrease in every column compared to the baseline setting.

Notably, in Table A.14, different patterns are observed for the relative fake fact detection improvement metric. While still decreasing on the final training benchmark, the cosine method improves on the non-final training benchmarks, up to 4,54 percentage points compared to the baseline. However, the no comparison model mostly only decreases compared to the baseline setting. This discrepancy between the metrics raises the question of whether the F1 score is a good metric for hallucination detection. Still, the cosine classifier does not yet improve the fake fact detection for the non-final training benchmarks on average.

**Selective Runs**

The tables discussed so far only yield the averages over all executed runs. In subsection A.3.4 the results regarding the selected benchmark orders are shown. The different experiments show that the highest relative fake fact improvements were attained with pretraining compared to the baseline setting of subsection 5.3.1. When training on HaluEval and then on TruthfulQA, the cosine model improved the relative fake fact metric by +15.68 percentage points. The maximum value achieved in subsection 5.3.1 was +13.17. In addition, this was the highest average test F1 score achieved on TruthfulQA in a total of 0.7. Similar things are shown with HaluEval, where the cosine model with pretraining on TruthfulQA and freezing contained the highest averaged relative fake fact improvement of +20.32 compared to the original maximum of +19.84. For ReFact, the maximum result reached +3.19 with the cosine model and pretraining, in contrast to the default of -1.42. Notably, even though not having an overall high performance, the runs that were pretrained on HaluEval and then trained on ReFact combined with freezing the aggregation module were the only ones increasing the fake fact detection accuracy over all benchmarks, visible in Table A.21. Still by no large margin. This shows that the idea of pretraining and freezing the aggregation module can increase performance in certain settings.

**Pretraining with/without Weight Freezing**

No large effect was measured comparing the scores achieved with pretraining and additionally freezing. In Table 5.8, the F1 scores only change minimally by a maximum margin of 0.02 between the columns, which stays the same for the relative fake fact improvement in Table A.14 with only a margin of 0.85.

**Reasons for Inconsistent Improvement**

Despite a few notable changes in performance, no consistent improvement was derived from the experimental approach. Two reasons for this were identified.

One reason might be that initializing the classifier with pretrained weights prevents the training of new meaningful representations for the other benchmark.

Especially considering that the works of [Bei+24; CH-+24; Ji+24; LH24] and subsection 6.2.1 showed that probing like architectures don't generalize to new benchmarks. Hypothetically, the weights of the feature encoder might be the major reason for this, as this module is designed to provide encodings. Therefore, future research might consider to only initialize the aggregation module with pretrained weights.

In addition, some training orders might not be practicable. An example is the experiment which pretrained on TruthfulQA and then trained on ReFact, where 29 of 40 experimental runs did not achieve a validation F1 score higher than zero (see subsection A.3.4). In contrast to that, only 11 runs were filtered out for all the other benchmark training orders. ReFact and TruthfulQA are both datasets with a sample size in the margin of 1000 (see section A.2). Therefore, the sample size might not be large enough to learn transferable weights. Considering the results of section 6.2, ReFact might be too complex to be used as a pretraining dataset. Due to these identified issues, future work should include additional benchmarks or discard impractical training orders.

**Answer RQ3**

Summarized, no large effect was measured with pretraining or freezing the aggregation module. However, in a few cases, the experiments showed improvements in generalization abilities and even achieved the overall best values on the test benchmarks, highlighting the potential of future research in this direction.

## 6.4 RQ4: How does the method compare to other hallucination detection methods that utilize the internal layers of LLMs?

In section 5.5, four baselines were derived from the related work. To compare them to the proposed methods, experiments regarding the model selection of section 5.2 and cross-benchmark evaluation subsection 5.3.1 were conducted. After describing the baselines briefly, the achieved results are discussed, and a conclusion is drawn.

**Baseline Groups**

The major difference between the four baselines is how they use the internal LLM layers as an input. Either by classifying solely on the last or middle layer, or by stacking all layers, or by employing an ensemble and classifying each layer individually. For an improved discussion, they were named as LastLayer, MiddleLayer, StackedLayers, and AllLayersEnsemble, respectively. To ensure comparability, these architectures all employed the same MLP architecture of the feature encoder module. As a result, they are no direct reproductions of the related work.

However, the experimental results are not used for a direct comparison to the related work. Instead, they are employed to evaluate whether the unique architecture features of the proposed method can increase the performance of internal hallucination detection methods. If so, the ideas of the other papers could be combined with the proposed methodology, like building on the different transformer block components, as done by [Bei+24].

**Model Selection**

To decide on their layer depth, all four baselines went through the same model selection experiments with HaluEval, visible in Table 5.9. With that, the models that achieved the maximum average of the 10 executed early stopping runs were used for further experiments. For the LastLayer model, the best F1 score of 0.78 was achieved with a layer depth of 2. For the MiddleLayer and the StackedLayers baselines, the maximum was attained with a classifier depth of 3, with values of 0.78 and 0.75, respectively. Lastly, the AllLayersEnsemble obtained its best value of 0.78 with a layer depth of one. With that, all maximum F1 scores averages achieved by the baselines were slightly worse compared to the proposed method, as the No Comparison achieved an averaged F1 score of 0.79 and the Cosine model one of 0.82 in Table 5.3.

**Non-Training Benchmarks**

After the model selection, the baselines were evaluated in the cross-benchmark evaluation, and their results on the non-training benchmark are discussed first. The Table 5.10 shows a large performance decrease of at least 0.17 F1 score percentage points between the training benchmark and the non-training benchmarks. This pattern is repetitive as shown in Table A.23, where the relative fake fact improvement is only affected negatively by the baselines on the unseen benchmarks. Similar results are visible for the proposed methods, which were already discussed in section 6.2 and section 6.3. A direct comparison between the models and the baselines shows no significant differences, as both achieved the maximum F1 score of 0.44. In the fake fact detection metric, the proposed models attained better values, still only by a minimal margin of 0.15 to the best achieved baseline value. However, as the cosine and no comparison both negatively decrease this metric on the non-training benchmarks, this has no significance.

**Training Benchmarks**

In contrast, the proposed methods outperform the baselines for every reported value calculated on all benchmarks and the training benchmarks for both metrics. Precisely, the minimum difference on the F1 Score is 0.01 for all benchmarks and 0.03 on the final training benchmark. Respectively, for the real fake fact improvement metrics, a margin of 0.13 for all benchmarks and one of 1.39 for the final training is achieved. As this is only the minimum, larger values exist. However, the baselines improved the relative fake fact improvement metric positively when tested solely on the training benchmark, with a maximum average of +13.87 employing the AllLayersEnsemble.

**Answer RQ4**

The main distinction of the proposed methodology to the baselines was how the internal layers of the LLMs were used as an input. The proposed method almost constantly obtained better values than the baselines. Therefore, the unique architectural features, like encoding every layer with the same MLP and comparing these encodings were proven as valuable for the domain. This encourages new research in combining the different methods of the related work with the proposed architecture, like using attention heads as done by [Bei+24].

To conclude, both the baselines and the proposed methods do not transfer well to benchmarks, not used for training. However, the proposed methods obtained better averages on the training benchmarks compared to all baselines and in the model selection experiments, highlighting the potential of the proposed architecture.

## 6.5 RQ5: How precisely can the proposed method identify hallucinated spans within the input text?

In section 5.6 the sequential labeling capabilities of the method were explored. Specifically, it was measured whether the selected methods, combined with tagging schemes and a CRF or greedy decoding, can identify the hallucinated spans provided by the ReFact benchmark. The obtained results are discussed to answer the final research question.

**Decoding with CRF**

The localization results employing a CRF reported in Table 5.11 are worse than the results achieved on the text classification results, with only achieving a maximum F1 score of 0.35 with no comparison and 0.34 with cosine. However, the other metrics of every run in Table A.29 show that all runs achieve a high accuracy between 95% and 96%. In general, the proposed methodology attains better results than the baselines. The mean F1 score of the 2 selected methods over all tagging schemes values in 0.28, while the mean over all baselines results in 0.18. Furthermore, the maximum achieved value of the baselines by the AllLayerEnsemble with 0.32 is worse than the maxima 0.35 with the no comparison and 0.34 with the cosine. By comparing the different tagging schemes, the BIO scheme performed best with an average of 0.27 compared to 0.18 for the IO scheme and 0.21 for the BIOES scheme.

**Greedy Decoding**

Without the CRF module, the trained classifiers decrease in performance. The reported results of Table A.28 only achieved an average F1 score of 0.03 and a maximum F1 Score of 0.15. In contrast to that, the runs employing a CRF benchmarked an average of 0.22. Notably, the runs without a CRF module obtained high accuracy, all ranging from 0.95 to 0.96, visible in Table A.30.

**Reasons for the modest F1 Scores**

Potential reasons for the modest F1 scores were identified.

First, a highly imbalanced label distribution is visible in Table A.3. In total, only around 4% of all tokens were labeled as positive, which corresponds to the maximum accuracy of 96%. With that, the main reason for the modest result might be due to the few existing data points. Methods addressing highly imbalanced datasets, like under-/overs-sampling or class weights, could potentially increase the performance of the method[Hai+17].

On the other hand, the spans might not correspond effectively to hallucinatory behavior. In the refact benchmark, whole sentences and keyphrases were altered and a focus was on coherently inducing these fake facts as plausible. Therefore, even arguably irrelevant subparts are marked positive, for example, introductory keyphrases starting a sentence. Future research might analyze the predicted tags of the proposed models and investigate whether the predictions indicate more insightful spans. Despite this, the fine-grained approach of ReFact, might be too complex for the selected training scenario. Similar results were already discussed in section 6.2 and section 6.3.

| RQ5: How precisely can the proposed method identify hallucinated spans within the input text? | Section 6.5 |

**Answer RQ5**

Summarized, despite outperforming the baselines and achieving a high accuracy of up to 96%, the method did not identify the marked hallucination spans precisely. Whether this is due to conceptual reasons or results from the high class imbalance remains an open question.

# 7 Conclusion

This thesis explored the potential of detecting hallucinations with the internal layers of LLMs on the benchmarks ReFact, HaluEval, and TruthfulQA. Specifically, a new architecture was proposed that leverages these internal layers as an input to classify hallucinated outputs of LLMs in a fake fact detection task. Afterward, its capabilities were evaluated in generalization tests, pretraining/freezing evaluations, and baseline comparisons. Additionally, the performance of the new model in sequential labeling was investigated.

**RQ1: How can a hallucination detection method be built that leverages the internal layers of LLMs?**

In this context, the first research question (RQ1) aimed to answer how a hallucination detection method can be built that leverages the internal layers of LLMs. A focus was set on designing a model that can adjust and compare the impact of different LLM layers. In total, 130 model variations were tested on HaluEval to detect hallucinations created by an LLM, which aimed to solve a fake fact detection task. A classifier that encodes every layer of an LLM with the same MLP was found to be capable of detecting hallucinations on the HaluEval benchmark to a certain margin. This classifier either compares these encodings with a cosine similarity or forwards without comparison and then aggregates these values to a final classification score with one linear layer. An average F1 score of 0.82 was obtained by the model that leverages the cosine similarity. Respectively, the model that did not compare internal encodings resulted in 0.79. These two methods were selected for further experimental exploration.

**RQ2: To what extent does the proposed method generalize across other benchmarks or to other LLMs?**

The second research question (RQ2) targeted whether the selected methods generalize across benchmarks or LLMs. Consequently, both were trained on each benchmark and tested across all benchmarks afterward.

First, their performance was measured on the test split of their training benchmark. On HaluEval, an average F1 score of 0.79 was achieved for both methods, followed by 0.66/0.65 on TruthfulQA (cosine / no comparison) and by 0.65 on ReFact.

Additionally, the accuracy of the fake fact detection task changes by +19.60/+19.84 (cosine / no comparison) on HaluEval, by +12.75/+13.17 on TruthfulQA, and by -3.01 on ReFact when employing the classifiers to detect LLM hallucinations. This decrease in accuracy and an identified training instability support the claims of the ReFact paper regarding the benchmarks complexity [Wan+25].

Tested on the other two benchmarks, the overall average F1 score results in only 0.43 and the models decrease the accuracy of the fake fact detection task on every benchmark. Therefore, the proposed methodology does not generalize effectively across benchmarks in the tested scenario. This aligns with prior research which reported similar generalization issues [Bei+24; CH-+24; Ji+24; LH24; Liu+23].

No major difference was observed between the methods, with an average F1 score of 0.72/0.71 (cosine / no comparison) on the training benchmark and an average of 0.44/0.45 on the benchmarks not used for training.

This is in contrast to the generalization test regarding different LLMs. By testing with 8 different LLMs on the benchmark TruthfulQA, notable differences between the two methods were revealed. The no comparison model did transfer well to other LLMs. As an average F1 score of 0.60 on the development LLM was attained, the overall performance does not decrease significantly compared to the average F1 score of 0.63 on all LLMs. No significant divergences between LLM size and model family were identified. However, this is not the case for the cosine method. The model achieved an average F1 score of 0.51 on all LLMs, compared to the 0.64 on the development LLM and the performance decreases for both LLM architecture families and steadily declines with increased LLM size. Label imbalance and insufficient sample sizes were identified as potential reasons for the decline of the cosine approach.

**RQ3: How do learned weight distributions corresponding to LLM layers affect the performance of the method combined with pretraining or freezing?**

To enhance performance or mitigate generalization issues, research question three (RQ3) analyzed learned weights and their potential when combined with pretraining or freezing.

As the proposed method trains weights associated with LLM layers in the aggregation module, these weights were plotted for the cosine and the no comparison method after one training on HaluEval. The resulting distributions show that larger weights correspond to the middle to later layers, supporting the findings of related work [Bei+24; CH-+24; Ji+24; SMZ24].

Afterward, experiments were conducted that successfully trained the selected methods on two benchmarks, with optionally freezing the weights of the aggregation module in the second training. To measure their effect on mitigating the generalization problem, they were tested in the cross-benchmark scenario of the previous research question (RQ2).

Neither idea consistently improved the average for both methods. The original cross-benchmark experiments achieved an average F1 score of 0.53 over all runs, while the pretraining experiments resulted in 0.51, and the ones freezing the aggregation module in 0.52. This pattern of decrease was maintained when tested solely on the training or non-training benchmarks, even for the relative fake fact improvement metric.

Only by comparing the methods separately, small differences were observed. The no comparison method averaged F1 scores improved slightly, calculated on all and non-final training benchmarks, while the cosine model's F1 scores only decreased. Notably this was different with the relative fake fact improvement metric, as here the cosine model improves on the non-final training benchmarks while the no comparison model decreases. In addition, no major difference between pretraining with or without freezing the aggregation module was revealed.

Still, particular benchmark training orders enabled the method to achieve its overall best performance in both metrics. As an example, by leveraging the cosine model with pretraining on HaluEval and weight freezing, the fake fact detection accuracy was improved on the training benchmark by + 15.68 for TruthfulQA and by +3.19 for ReFact. When pretrained on TruthfulQA and then HaluEval without freezing, the metric was improved by +20.32 on HaluEval. Notably, only by training the cosine model with freezing on HaluEval, then ReFact, fake fact detection accuracy was improved on all benchmarks, still by no large margin. This shows that the idea of pretraining and freezing the aggregation module can increase performance in certain settings. Still, the idea did not improve the performance consistently.

**RQ4: How does the method compare to other hallucination detection methods that utilize the internal layers of LLMs?**

The fourth research question (RQ4) compared baselines that classify internal layers with the proposed methods. Four baselines were derived from the related work, which differ in the way they leverage the internal layers as input. Either by classifying solely on the last or middle layer, or by stacking all layers, or by employing an ensemble and classifying each layer individually. For an improved comparison, they were all implemented with the same MLP of the feature extractor proposed in the methodology. Therefore, they better reflect whether the idea of encoding every layer with the same MLP, comparing, and aggregating can improve performance. The baselines and methods were compared in experiments regarding the model selection of (RQ1) and cross-benchmark evaluation of (RQ2).

All baselines and the two proposed models did not transfer well to benchmarks, not utilized for training. However, the proposed methods consistently obtained better values than the baselines on all training benchmarks, highlighting the capabilities of the proposed architecture.

**RQ5: How precisely can the proposed method identify hallucinated spans within the input text?**

Finally, (RQ5) aimed to test whether the proposed method can identify hallucinated spans within the input text provided by the ReFact benchmark. This problem was formulated as a sequential labeling task by incorporating tagging schemes. IO, BIO and BIOES were explored and combined with either greedy decoding or employing a conditional random field (CRF) on the emission scores of the proposed models and baselines.

After one training run for every model variation, the metrics F1 score and accuracy were discussed. Compared to the baselines, the selected models almost constantly attained better F1 scores, supporting the conclusion of (RQ4). On average, the results of the CRF and BIO tagging performed best. However, the highest F1 score was only 0.35 obtained by the no comparison model in combination with a CRF and BIO tagging. Without the CRF module, the trained classifiers decrease in performance a maximum F1 score of 0.15, still achieving similar accuracy. Notably, all runs achieved a high accuracy between 95% and 96%.

The discussion identified multiple reasons. First, ReFact provides a highly imbalanced label distribution as only 4% of all tokens were labeled as positive. On the other hand, the spans might not correspond effectively to hallucinatory behavior or the benchmark might be to complex for the selected training scenario. In the end, whether the modest results are due to conceptual reasons or high class imbalance remains an open question.

### 7.0.1 Limitations

To feasibly explore the topic of hallucination detection with the internal layers of LLMs, the scope had to be reduced. Only instruction fine-tuned open source LLMs based on the decoder part of the transformer were discussed in this thesis and other LLM variations were ignored. In total, this resulted in only 8 LLMs of two model families. Similarly, with no focus on other tasks or even other domains, only three benchmarks were employed, all being discriminatory factual question answering datasets. Additionally, only a small range of model architectures was targeted in this thesis, with only one training pipeline and one selected hyperparameter setting. Despite that, only two metrics were mainly explored. As section 6.3 showed a discrepancy between the F1 score and the relative fake fact improvement, an additional question arises whether both metrics are applicable for hallucination. With that, all concluded answers are reduced to this subset of selections.

Design choices of the methodology were influenced by related work. Despite a critical experimental approach, the methodology is founded on the reliability of these reports. Due to the unavailability of publicly available code, the direct comparison to the related work was discarded. Therefore, only self-implemented baselines were investigated. Additionally, for the task of sequential labeling, no direct hallucinations of LLMs were explored, as the labeled ground truth was derived from the benchmark, and tagging hallucinated generations was considered as impractical. Despite that, the proposed methodology is a white box approach and is limited to accessible internal layers of LLMs. Finally, the generalization issues raise concerns about whether probing-based methods should be employed in general.

### 7.0.2 Outlook

Nonetheless, this work identified multiple potential research directions that might mitigate the generalization issues or could increase performance.

In (RQ1), only limited model variations were discussed. Therefore, future research might enhance this by exploring different encoders, comparison functions, or changing the hidden sizes or activation functions of the current MLPs. Furthermore, even altering the training pipeline with different hyperparameters or loss functions could be impactful.

(RQ2) revealed generalization issues and decreased training stability of probing-based methods, again motivating adjustments in future work on the selected training scenario or selected architecture.

The explored weight distributions of (RQ3) showed the impact of middle to later LLM layers. A new potential research direction might evaluate whether dropping the earlier layers or preselected distributions could increase the classifier's performance. Furthermore, the topic of pretraining and weight freezing was analyzed. While not consistently, selective runs improved performance and mitigated the generalization issues slightly. With that, future work should include additional benchmarks or discard impractical training orders.

Results of (RQ4) demonstrated the ability of the method to outperform self-implemented baselines. This encourages new research in combining the different methods of the related work with the proposed architecture, like using attention heads as done by [Bei+24].

For the modest performance of the sequential labeling in (RQ5), issues with the training data were identified. A high label imbalance motivates, enhancing the training pipeline with methods that address this, like under- and oversampling or class weights (see [Hai+17]). Additionally, analyzing the predicted spans of the proposed methods might reveal insights, as the labeled spans might not correspond to hallucinations.

Addressing these future research directions might improve the prediction performance or mitigate the generalization issues and, consequently, reduce the potential harm of hallucinations on society.

# Bibliography

[Agr+24]   Ayush Agrawal, Mirac Suzgun, Lester Mackey, and Adam Tauman Kalai. *Do Language Models Know When They're Hallucinating References?* 2024. arXiv: 2305.18248 [cs.CL]. URL: https://arxiv.org/abs/2305.18248 (see page 13).

[Ahd+24]   Gustaf Ahdritz, Tian Qin, Nikhil Vyas, Boaz Barak, and Benjamin L. Edelman. *Distinguishing the Knowable from the Unknowable with Language Models.* 2024. arXiv: 2402.03563 [cs.LG]. URL: https://arxiv.org/abs/2402.03563 (see pages 1, 14, 32).

[AM23]    Amos Azaria and Tom Mitchell. *The Internal State of an LLM Knows When It's Lying.* 2023. arXiv: 2304.13734 [cs.CL]. URL: https://arxiv.org/abs/2304.13734 (see pages ii, iii, 1, 14, 20, 29).

[Bei+24]   Mohammad Beigi, Ying Shen, Runing Yang, Zihao Lin, Qifan Wang, Ankith Mohan, Jianfeng He, Ming Jin, Chang-Tien Lu, and Lifu Huang. *InternalInspector $I^2$: Robust Confidence Estimation in LLMs through Internal States.* 2024. arXiv: 2406.12053 [cs.CL]. URL: https://arxiv.org/abs/2406.12053 (see pages ii, iii, 1, 2, 14, 15, 19, 29, 33, 39, 41, 43–45, 49, 53).

[Bie20]   Lukas Biewald. *Experiment Tracking with Weights and Biases.* Software available from wandb.com. 2020. URL: https://www.wandb.com/ (see page 18).

[Bur+24]   Collin Burns, Haotian Ye, Dan Klein, and Jacob Steinhardt. *Discovering Latent Knowledge in Language Models Without Supervision.* 2024. arXiv: 2212.03827 [cs.CL]. URL: https://arxiv.org/abs/2212.03827 (see pages ii, iii, 1, 14, 15).

[CH-+24]   Sky CH-Wang, Benjamin Van Durme, Jason Eisner, and Chris Kedzie. *Do Androids Know They're Only Dreaming of Electric Sheep?* 2024. arXiv: 2312.17249 [cs.CL]. URL: https://arxiv.org/abs/2312.17249 (see pages 2, 15, 16, 19, 21, 33, 39, 41, 43, 49).

[Che+23a]  Anthony Chen, Panupong Pasupat, Sameer Singh, Hongrae Lee, and Kelvin Guu. *PURR: Efficiently Editing Language Model Hallucinations by Denoising Language Model Corruptions.* 2023. arXiv: 2305.14908 [cs.CL]. URL: https://arxiv.org/abs/2305.14908 (see page 16).

[Che+23b]  I-Chun Chern, Steffi Chern, Shiqi Chen, Weizhe Yuan, Kehua Feng, Chunting Zhou, Junxian He, Graham Neubig, and Pengfei Liu. *FacTool: Factuality Detection in Generative AI – A Tool Augmented Framework for Multi-Task and Multi-Domain Scenarios.* 2023. arXiv: 2307.13528 [cs.CL]. URL: https://arxiv.org/abs/2307.13528 (see page 12).

[Che+24]   Chao Chen, Kai Liu, Ze Chen, Yi Gu, Yue Wu, Mingyuan Tao, Zhihang Fu, and Jieping Ye. *INSIDE: LLMs' Internal States Retain the Power of Hallucination Detection*. 2024. arXiv: 2402.03744 [cs.CL]. URL: https://arxiv.org/abs/2402.03744 (see page 15).

[Chu+24]   Yung-Sung Chuang, Yujia Xie, Hongyin Luo, Yoon Kim, James Glass, and Pengcheng He. *DoLa: Decoding by Contrasting Layers Improves Factuality in Large Language Models*. 2024. arXiv: 2309.03883 [cs.CL]. URL: https://arxiv.org/abs/2309.03883 (see page 2).

[CRG23]    James Campbell, Richard Ren, and Phillip Guo. *Localizing Lying in Llama: Understanding Instructed Dishonesty on True-False Questions Through Prompting, Probing, and Patching*. 2023. arXiv: 2311.15131 [cs.LG]. URL: https://arxiv.org/abs/2311.15131 (see pages 1, 14).

[DFO20]    Marc Peter Deisenroth, A. Aldo Faisal, and Cheng Soon Ong. *Mathematics for Machine Learning*. Cambridge University Press, 2020. URL: https://mml-book.com (see pages 10, 11, 20).

[Dhu+23]   Shehzaad Dhuliawala, Mojtaba Komeili, Jing Xu, Roberta Raileanu, Xian Li, Asli Celikyilmaz, and Jason Weston. *Chain-of-Verification Reduces Hallucination in Large Language Models*. 2023. arXiv: 2309.11495 [cs.CL]. URL: https://arxiv.org/abs/2309.11495 (see page 12).

[DYT24]    Hanyu Duan, Yi Yang, and Kar Yan Tam. *Do LLMs Know about Hallucination? An Empirical Investigation of LLM's Hidden States*. 2024. arXiv: 2402.09733 [cs.CL]. URL: https://arxiv.org/abs/2402.09733 (see pages ii, iii, 1, 14).

[Fad+24]   Ekaterina Fadeeva, Aleksandr Rubashevskii, Artem Shelmanov, Sergey Petrakov, Haonan Li, Hamdy Mubarak, Evgenii Tsymbalov, Gleb Kuzmin, Alexander Panchenko, Timothy Baldwin, Preslav Nakov, and Maxim Panov. *Fact-Checking the Output of Large Language Models via Token-Level Uncertainty Quantification*. 2024. arXiv: 2403.04696 [cs.CL]. URL: https://arxiv.org/abs/2403.04696 (see page 16).

[Gao+23]   Luyu Gao, Zhuyun Dai, Panupong Pasupat, Anthony Chen, Arun Tejasvi Chaganty, Yicheng Fan, Vincent Y. Zhao, Ni Lao, Hongrae Lee, Da-Cheng Juan, and Kelvin Guu. *RARR: Researching and Revising What Language Models Say, Using Language Models*. 2023. arXiv: 2210.08726 [cs.CL]. URL: https://arxiv.org/abs/2210.08726 (see page 16).

[Gaw+22]   Jakob Gawlikowski, Cedrique Rovile Njieutcheu Tassi, Mohsin Ali, Jongseok Lee, Matthias Humt, Jianxiang Feng, Anna Kruspe, Rudolph Triebel, Peter Jung, Ribana Roscher, Muhammad Shahzad, Wen Yang, Richard Bamler, and Xiao Xiang Zhu. *A Survey of Uncertainty in Deep Neural Networks*. 2022. arXiv: 2107.03342 [cs.LG]. URL: https://arxiv.org/abs/2107.03342 (see pages 13, 21).

[GBC16]   Ian Goodfellow, Yoshua Bengio, and Aaron Courville. **Deep Learning**. http://www.deeplearningbook.org. MIT Press, 2016 (see pages 9, 10, 23).

[GDJ24]   Aaron Grattafiori, Abhimanyu Dubey, and Abhinav Jauhri. *The Llama 3 Herd of Models*. 2024. arXiv: 2407.21783 [cs.AI]. URL: https://arxiv.org/abs/2407.21783 (see pages 26, 30).

[Hai+17]  Guo Haixiang, Li Yijing, Jennifer Shang, Gu Mingyun, Huang Yuanyue, and Gong Bing. **Learning from class-imbalanced data: Review of methods and applications**. *Expert Systems with Applications* 73 (2017), 220–239. ISSN: 0957-4174. DOI: https://doi.org/10.1016/j.eswa.2016.12.035. URL: https://www.sciencedirect.com/science/article/pii/S0957417416307175 (see pages 46, 53).

[He+20]   Zhiyong He, Zanbo Wang, Wei Wei, Shanshan Feng, Xianling Mao, and Sheng Jiang. *A Survey on Recent Advances in Sequence Labeling from Deep Learning Models*. 2020. arXiv: 2011.06727 [cs.CL]. URL: https://arxiv.org/abs/2011.06727 (see pages 8, 9, 26).

[Hua+24]  Lei Huang, Weijiang Yu, Weitao Ma, Weihong Zhong, Zhangyin Feng, Haotian Wang, Qianglong Chen, Weihua Peng, Xiaocheng Feng, Bing Qin, and Ting Liu. **A Survey on Hallucination in Large Language Models: Principles, Taxonomy, Challenges, and Open Questions**. *ACM Transactions on Information Systems* (Nov. 2024). ISSN: 1558-2868. DOI: 10.1145/3703155. URL: http://dx.doi.org/10.1145/3703155 (see pages ii, iii, 1, 7, 8, 12, 13, 16).

[Hua+25]  Yuheng Huang, Jiayang Song, Zhijie Wang, Shengming Zhao, Huaming Chen, Felix Juefei-Xu, and Lei Ma. **Look Before You Leap: An Exploratory Study of Uncertainty Analysis for Large Language Models**. *IEEE Transactions on Software Engineering* 51:2 (Feb. 2025), 413–429. ISSN: 2326-3881. DOI: 10.1109/tse.2024.3519464. URL: http://dx.doi.org/10.1109/TSE.2024.3519464 (see page 13).

[IAR23]   Mohammadreza Iman, Hamid Reza Arabnia, and Khaled Rasheed. **A Review of Deep Transfer Learning and Recent Advancements**. *Technologies* 11:2 (Mar. 2023), 40. ISSN: 2227-7080. DOI: 10.3390/technologies11020040. URL: http://dx.doi.org/10.3390/technologies11020040 (see page 23).

[Ji+23]   Ziwei Ji, Nayeon Lee, Rita Frieske, Tiezheng Yu, Dan Su, Yan Xu, Etsuko Ishii, Ye Jin Bang, Andrea Madotto, and Pascale Fung. **Survey of Hallucination in Natural Language Generation**. *ACM Comput. Surv.* 55:12 (Mar. 2023). ISSN: 0360-0300. DOI: 10.1145/3571730. URL: https://doi.org/10.1145/3571730 (see pages 8, 16).

[Ji+24] Ziwei Ji, Delong Chen, Etsuko Ishii, Samuel Cahyawijaya, Yejin Bang, Bryan Wilie, and Pascale Fung. *LLM Internal States Reveal Hallucination Risk Faced With a Query*. 2024. arXiv: 2407.03282 [cs.CL]. URL: https://arxiv.org/abs/2407.03282 (see pages ii, iii, 1, 2, 14, 15, 19, 32, 39, 41, 43, 49).

[JM25] Daniel Jurafsky and James H. Martin. *Speech and Language Processing: An Introduction to Natural Language Processing, Computational Linguistics, and Speech Recognition with Language Models*. 3rd. Online manuscript released January 12, 2025. 2025. URL: https://web.stanford.edu/~jurafsky/slp3/ (see pages 5–10, 18, 22, 23, 26).

[Kad+22] Saurav Kadavath, Tom Conerly, Amanda Askell, Tom Henighan, Dawn Drain, Ethan Perez, Nicholas Schiefer, Zac Hatfield-Dodds, Nova DasSarma, Eli Tran-Johnson, Scott Johnston, Sheer El-Showk, Andy Jones, Nelson Elhage, Tristan Hume, Anna Chen, Yuntao Bai, Sam Bowman, Stanislav Fort, Deep Ganguli, Danny Hernandez, Josh Jacobson, Jackson Kernion, Shauna Kravec, Liane Lovitt, Kamal Ndousse, Catherine Olsson, Sam Ringer, Dario Amodei, Tom Brown, Jack Clark, Nicholas Joseph, Ben Mann, Sam McCandlish, Chris Olah, and Jared Kaplan. *Language Models (Mostly) Know What They Know*. 2022. arXiv: 2207.05221 [cs.CL]. URL: https://arxiv.org/abs/2207.05221 (see pages 13, 17, 18).

[Kay23] Byron Kaye. *Australian mayor readies world's first defamation lawsuit over ChatGPT content*. Reuters (Apr. 2023). URL: https://www.reuters.com/technology/australian-mayor-readies-worlds-first-defamation-lawsuit-over-chatgpt-content-2023-04-05/ (see pages ii, iii, 1).

[Kur18] Kemal Kurniawan. *pytorch-crf: Conditional Random Field module for PyTorch*. Accessed: 2025-06-02. 2018. URL: https://pytorch-crf.readthedocs.io/en/stable/index.html (see page 22).

[LH24] Benjamin A. Levinstein and Daniel A. Herrmann. *Still no lie detector for language models: probing empirical and conceptual roadblocks*. Philosophical Studies (Feb. 2024). ISSN: 1573-0883. DOI: 10.1007/s11098-023-02094-3. URL: http://dx.doi.org/10.1007/s11098-023-02094-3 (see pages 2, 15, 39, 43, 49).

[LHE22a] Stephanie Lin, Jacob Hilton, and Owain Evans. *Teaching Models to Express Their Uncertainty in Words*. 2022. arXiv: 2205.14334 [cs.CL]. URL: https://arxiv.org/abs/2205.14334 (see page 13).

[LHE22b] Stephanie Lin, Jacob Hilton, and Owain Evans. *TruthfulQA: Measuring How Models Mimic Human Falsehoods*. 2022. arXiv: 2109.07958 [cs.CL]. URL: https://arxiv.org/abs/2109.07958 (see pages 2, 12, 24).

[Li+23]   Junyi Li, Xiaoxue Cheng, Wayne Xin Zhao, Jian-Yun Nie, and Ji-Rong Wen. *HaluEval: A Large-Scale Hallucination Evaluation Benchmark for Large Language Models*. 2023. arXiv: 2305.11747 [cs.CL]. URL: https://arxiv.org/abs/2305.11747 (see pages 2, 24).

[Li+24a]  Junyi Li, Jie Chen, Ruiyang Ren, Xiaoxue Cheng, Wayne Xin Zhao, Jian-Yun Nie, and Ji-Rong Wen. *The Dawn After the Dark: An Empirical Study on Factuality Hallucination in Large Language Models*. 2024. arXiv: 2401.03205 [cs.CL]. URL: https://arxiv.org/abs/2401.03205 (see page 12).

[Li+24b]  Kenneth Li, Oam Patel, Fernanda Viégas, Hanspeter Pfister, and Martin Wattenberg. *Inference-Time Intervention: Eliciting Truthful Answers from a Language Model*. 2024. arXiv: 2306.03341 [cs.LG]. URL: https://arxiv.org/abs/2306.03341 (see pages 1, 14, 15).

[Liu+22]  Tianyu Liu, Yizhe Zhang, Chris Brockett, Yi Mao, Zhifang Sui, Weizhu Chen, and Bill Dolan. *A Token-level Reference-free Hallucination Detection Benchmark for Free-form Text Generation*. 2022. arXiv: 2104.08704 [cs.CL]. URL: https://arxiv.org/abs/2104.08704 (see pages 12, 16).

[Liu+23]  Kevin Liu, Stephen Casper, Dylan Hadfield-Menell, and Jacob Andreas. *Cognitive Dissonance: Why Do Language Model Outputs Disagree with Internal Representations of Truthfulness?* 2023. arXiv: 2312.03729 [cs.CL]. URL: https://arxiv.org/abs/2312.03729 (see pages 15, 39, 49).

[LM24]    Grant Ledger and Rafael Mancinni. *Detecting LLM Hallucinations Using Monte Carlo Simulations on Token Probabilities*. June 2024. DOI: 10.36227/techrxiv.171822396.61518693/v1 (see page 16).

[LS24]    Haoyan Luo and Lucia Specia. *From Understanding to Utilization: A Survey on Explainability for Large Language Models*. 2024. arXiv: 2401.12874 [cs.CL]. URL: https://arxiv.org/abs/2401.12874 (see page 10).

[Min+23]  Sewon Min, Kalpesh Krishna, Xinxi Lyu, Mike Lewis, Wen-tau Yih, Pang Wei Koh, Mohit Iyyer, Luke Zettlemoyer, and Hannaneh Hajishirzi. *FActScore: Fine-grained Atomic Evaluation of Factual Precision in Long Form Text Generation*. 2023. arXiv: 2305.14251 [cs.CL]. URL: https://arxiv.org/abs/2305.14251 (see pages 12, 16).

[Min+25]  Shervin Minaee, Tomas Mikolov, Narjes Nikzad, Meysam Chenaghlu, Richard Socher, Xavier Amatriain, and Jianfeng Gao. *Large Language Models: A Survey*. 2025. arXiv: 2402.06196 [cs.CL]. URL: https://arxiv.org/abs/2402.06196 (see page 5).

[Mis+24]   Abhika Mishra, Akari Asai, Vidhisha Balachandran, Yizhong Wang, Graham Neubig, Yulia Tsvetkov, and Hannaneh Hajishirzi. *Fine-grained Hallucination Detection and Editing for Language Models*. 2024. arXiv: 2401.06855 [cs.CL]. URL: https://arxiv.org/abs/2401.06855 (see page 16).

[MLG23]   Potsawee Manakul, Adian Liusie, and Mark J. F. Gales. *SelfCheckGPT: Zero-Resource Black-Box Hallucination Detection for Generative Large Language Models*. 2023. arXiv: 2303.08896 [cs.CL]. URL: https://arxiv.org/abs/2303.08896 (see pages 13, 16).

[MT24]   Samuel Marks and Max Tegmark. *The Geometry of Truth: Emergent Linear Structure in Large Language Model Representations of True/False Datasets*. 2024. arXiv: 2310.06824 [cs.AI]. URL: https://arxiv.org/abs/2310.06824 (see pages 1, 14, 32).

[NH10]   Vinod Nair and Geoffrey E. Hinton. Rectified linear units improve restricted boltzmann machines. In: *Proceedings of the 27th International Conference on International Conference on Machine Learning*. ICML'10. Haifa, Israel: Omnipress, 2010, 807–814. ISBN: 9781605589077 (see page 20).

[Pas+19]   Adam Paszke, Sam Gross, Francisco Massa, Adam Lerer, James Bradbury, Gregory Chanan, Trevor Killeen, Zeming Lin, Natalia Gimelshein, Luca Antiga, Alban Desmaison, Andreas Köpf, Edward Yang, Zach DeVito, Martin Raison, Alykhan Tejani, Sasank Chilamkurthy, Benoit Steiner, Lu Fang, Junjie Bai, and Soumith Chintala. *PyTorch: An Imperative Style, High-Performance Deep Learning Library*. 2019. arXiv: 1912.01703 [cs.LG]. URL: https://arxiv.org/abs/1912.01703 (see pages 17, 22, 23).

[Ped+11]   F. Pedregosa, G. Varoquaux, A. Gramfort, V. Michel, B. Thirion, O. Grisel, M. Blondel, P. Prettenhofer, R. Weiss, V. Dubourg, J. Vanderplas, A. Passos, D. Cournapeau, M. Brucher, M. Perrot, and E. Duchesnay. Scikit-learn: Machine Learning in Python. *Journal of Machine Learning Research* 12 (2011), 2825–2830 (see page 18).

[Rud17]   Sebastian Ruder. *An overview of gradient descent optimization algorithms*. 2017. arXiv: 1609.04747 [cs.LG]. URL: https://arxiv.org/abs/1609.04747 (see page 9).

[Rum+24]   Prathiksha Rumale, Simran Tiwari, Tejas G. Naik, Sahil Gupta, Dung Ngoc Thai, Wenlong Zhao, Sunjae Kwon, Victor Ardulov, Karim Tarabishy, Andrew McCallum, and Wael Salloum. Faithfulness Hallucination Detection in Healthcare AI. In: ACM, 2024. URL: https://openreview.net/pdf?id=6eMIzKFOpJ (see pages ii, iii, 1).

[SMZ24]    Ben Snyder, Marius Moisescu, and Muhammad Bilal Zafar. *On Early Detection of Hallucinations in Factual Question Answering*. In: *Proceedings of the 30th ACM SIGKDD Conference on Knowledge Discovery and Data Mining*. KDD '24. ACM, Aug. 2024, 2721–2732. DOI: 10.1145/3637528.3671796. URL: http://dx.doi.org/10.1145/3637528.3671796 (see pages ii, iii, 1, 2, 14, 15, 19, 32, 41, 49).

[Su+24]    Weihang Su, Changyue Wang, Qingyao Ai, Yiran HU, Zhijing Wu, Yujia Zhou, and Yiqun Liu. *Unsupervised Real-Time Hallucination Detection based on the Internal States of Large Language Models*. 2024. arXiv: 2403.06448 [cs.CL]. URL: https://arxiv.org/abs/2403.06448 (see pages ii, iii, 2, 14, 15, 20, 32).

[tea20]    The pandas development team. *pandas-dev/pandas: Pandas*. Version latest. Feb. 2020. DOI: 10.5281/zenodo.3509134. URL: https://doi.org/10.5281/zenodo.3509134 (see page 18).

[Tea25]    Gemma Team. *Gemma 3* (2025). URL: https://goo.gle/Gemma3Report (see pages 26, 30).

[Vas+23]   Ashish Vaswani, Noam Shazeer, Niki Parmar, Jakob Uszkoreit, Llion Jones, Aidan N. Gomez, Lukasz Kaiser, and Illia Polosukhin. *Attention Is All You Need*. 2023. arXiv: 1706.03762 [cs.CL]. URL: https://arxiv.org/abs/1706.03762 (see page 5).

[Wan+25]   Yindong Wang, Martin Preiß, Jan Hoffbauer, Abdullatif Ghajar, Margarita Bugueño, and Gerard de Melo. *ReFACT: A Benchmark Based on r/AskScience for Misinformation Detection with LLMs*. Submitted to EMNLP 2025. 2025 (see pages 1, 2, 7, 8, 13, 16–18, 24, 38, 49).

[Wol+20]   Thomas Wolf, Lysandre Debut, Victor Sanh, Julien Chaumond, Clement Delangue, Anthony Moi, Pierric Cistac, Tim Rault, Rémi Louf, Morgan Funtowicz, Joe Davison, Sam Shleifer, Patrick von Platen, Clara Ma, Yacine Jernite, Julien Plu, Canwen Xu, Teven Le Scao, Sylvain Gugger, Mariama Drame, Quentin Lhoest, and Alexander M. Rush. *HuggingFace's Transformers: State-of-the-art Natural Language Processing*. 2020. arXiv: 1910.03771 [cs.CL]. URL: https://arxiv.org/abs/1910.03771 (see page 18).

[Wu+24]    Xuansheng Wu, Haiyan Zhao, Yaochen Zhu, Yucheng Shi, Fan Yang, Tianming Liu, Xiaoming Zhai, Wenlin Yao, Jundong Li, Mengnan Du, and Ninghao Liu. *Usable XAI: 10 Strategies Towards Exploiting Explainability in the LLM Era*. 2024. arXiv: 2403.08946 [cs.LG]. URL: https://arxiv.org/abs/2403.08946 (see page 10).

| | |
|---|---|
| [Xio+24] | Miao Xiong, Zhiyuan Hu, Xinyang Lu, Yifei Li, Jie Fu, Junxian He, and Bryan Hooi. *Can LLMs Express Their Uncertainty? An Empirical Evaluation of Confidence Elicitation in LLMs*. 2024. arXiv: 2306.13063 [cs.CL]. URL: https://arxiv.org/abs/2306.13063 (see page 13). |
| [Yad19] | Omry Yadan. *Hydra - A framework for elegantly configuring complex applications*. Github. 2019. URL: https://github.com/facebookresearch/hydra (see page 18). |
| [Yan+18] | Zhilin Yang, Peng Qi, Saizheng Zhang, Yoshua Bengio, William W. Cohen, Ruslan Salakhutdinov, and Christopher D. Manning. *HotpotQA: A Dataset for Diverse, Explainable Multi-hop Question Answering*. 2018. arXiv: 1809.09600 [cs.CL]. URL: https://arxiv.org/abs/1809.09600 (see page 24). |
| [Zak21] | Kevin Zakka. *A Playground for CLIP-like Models*. Version 0.0.1. July 2021. URL: https://github.com/kevinzakka/clip_playground (see page 18). |
| [Zha+23a] | Aston Zhang, Zachary C. Lipton, Mu Li, and Alexander J. Smola. **Dive into Deep Learning**. https://D2L.ai. Cambridge University Press, 2023 (see pages 5, 9, 10, 20, 23). |
| [Zha+23b] | Yue Zhang, Yafu Li, Leyang Cui, Deng Cai, Lemao Liu, Tingchen Fu, Xinting Huang, Enbo Zhao, Yu Zhang, Yulong Chen, Longyue Wang, Anh Tuan Luu, Wei Bi, Freda Shi, and Shuming Shi. *Siren's Song in the AI Ocean: A Survey on Hallucination in Large Language Models*. 2023. arXiv: 2309.01219 [cs.CL]. URL: https://arxiv.org/abs/2309.01219 (see pages 6–8, 16). |
| [Zha+25] | Wayne Xin Zhao, Kun Zhou, Junyi Li, Tianyi Tang, Xiaolei Wang, Yupeng Hou, Yingqian Min, Beichen Zhang, Junjie Zhang, Zican Dong, Yifan Du, Chen Yang, Yushuo Chen, Zhipeng Chen, Jinhao Jiang, Ruiyang Ren, Yifan Li, Xinyu Tang, Zikang Liu, Peiyu Liu, Jian-Yun Nie, and Ji-Rong Wen. *A Survey of Large Language Models*. 2025. arXiv: 2303.18223 [cs.CL]. URL: https://arxiv.org/abs/2303.18223 (see pages ii, iii, 1, 5). |
| [Zho+22] | Shaowen Zhou, Bowen Yu, Aixin Sun, Cheng Long, Jingyang Li, Haiyang Yu, Jian Sun, and Yongbin Li. *A Survey on Neural Open Information Extraction: Current Status and Future Directions*. 2022. arXiv: 2205.11725 [cs.CL]. URL: https://arxiv.org/abs/2205.11725 (see page 9). |

# Declaration of Authorship

I hereby declare that this thesis is my own unaided work. All direct or indirect sources used are acknowledged as references.

Potsdam, June 10, 2025  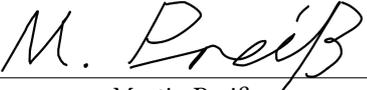
                        Martin Preiß

# A. Appendix

## A.1 Fake Fact Detection Prompt

**Fake Fact Detection Prompt**

```
Your task is to evaluate the factual correctness of a
given answer to a question.
Provide only a final verdict of either [TRUE] if
the entire answer is factually correct, or [FALSE]
if any part of the answer contains inaccuracies or
hallucinations.
Only output the Final Verdict.  No Explanation.

Question:  {question}
Answer:  {answer}
Final Verdict:  The answer is [
```

## A.2 Dataset Statistics

| Benchmark | Size | Percentage of Positives | | |
| --- | --- | --- | --- | --- |
| | | Train | Val | Test |
| ReFact | 2002 | 46.51 | 46.30 | 48.49 |
| HaluEval | 20000 | 35.20 | 35.50 | 33.73 |
| TruthfulQA | 1634 | 35.03 | 31.30 | 36.59 |

**Table A.1:** Text classification dataset sizes and the distribution of positives regarding the development LLM LLaMa-3.1-8B-Instruct for different benchmarks

| LLM | Layers | Embedding Size | Ratio of Positives | | |
|---|---|---|---|---|---|
| | | | Train | Val | Test |
| LLaMa-3.1-8B-Instruct | 32 | 4096 | 35.03 | 31.30 | 36.59 |
| Gemma-3-1b-it | 26 | 1152 | 49.91 | 51.22 | 50.81 |
| Gemma-3-4b-it | 34 | 2560 | 35.90 | 36.18 | 36.59 |
| Gemma-3-12b-it | 48 | 3840 | 22.59 | 17.48 | 21.95 |
| Gemma-3-27b-it | 62 | 5376 | 22.94 | 19.51 | 25.61 |
| LLaMa-3.2-1B-Instruct | 16 | 2048 | 52.28 | 54.07 | 56.10 |
| LLaMa-3.2-3B-Instruct | 28 | 3072 | 38.18 | 36.99 | 41.46 |
| LLaMa-3.3-70B-Instruct | 80 | 8192 | 20.49 | 17.48 | 23.17 |

**Table A.2:** Number of layers and embedding sizes of the selected LLMs with the distribution of positives regarding the TruthfulQA benchmark for the text classification task

| Split | Train | Val | Test | Total |
|---|---|---|---|---|
| Tokens | 222954 | 56139 | 55207 | 334300 |
| Positives (%) | 4.31 | 4.19 | 4.40 | 4.31 |

**Table A.3:** Token counts and positive label ratios for 2002 sequences(True/False) of the ReFact benchmark regarding the sequence labeling

## A.3 Experiments

### A.3.1 Model Selection

| Comparison Method | Layer Depth | | | | |
|---|---|---|---|---|---|
| | 1 | 2 | 3 | 4 | 5 |
| No Comparison | 0.56 | 0.63 | 0.08 | 0.00 | 0.00 |
| Pairwise Dot Product | 0.47 | 0.32 | 0.00 | 0.00 | 0.00 |
| Euclidean Distance | 0.71 | 0.49 | 0.39 | 0.33 | 0.08 |
| Manhattan Distance | 0.16 | 0.41 | 0.24 | 0.16 | 0.00 |
| Cosine Similarity | 0.72 | 0.49 | 0.24 | 0.00 | 0.00 |

**Table A.4:** F1-Scores measured with the HaluEval TestSplit of Different Comparison Methods and Layer Depths for the Shared Classifier Ensemble Aggregation (Non-Linear)

| Comparison Method | Layer Depth | | | | |
|---|---|---|---|---|---|
| | 1 | 2 | 3 | 4 | 5 |
| No Comparison | 0.72 | 0.63 | 0.32 | 0.23 | 0.00 |
| Pairwise Dot Product | 0.32 | 0.40 | 0.16 | 0.00 | 0.00 |
| Euclidean Distance | 0.53 | 0.64 | 0.80 | 0.31 | 0.08 |
| Manhattan Distance | 0.29 | 0.65 | 0.39 | 0.56 | 0.24 |
| Cosine Similarity | 0.71 | 0.56 | 0.40 | 0.00 | 0.00 |

**Table A.5:** F1-Scores with the HaluEval TestSplit of Different Comparison Methods and Layer Depths for the Shared Classifier Ensemble Aggregation (Linear)

| Comparison Method | Layer Depth | | | | |
|---|---|---|---|---|---|
| | 1 | 2 | 3 | 4 | 5 |
| No Comparison | 0.80 | 0.79 | 0.54 | 0.00 | 0.00 |
| Dot Product | 0.78 | 0.80 | 0.41 | 0.24 | 0.00 |
| Euclidean Norm | 0.71 | 0.80 | 0.80 | 0.15 | 0.08 |
| Manhattan Norm | 0.72 | 0.80 | 0.73 | 0.48 | 0.16 |
| Pairwise Dot Product | 0.71 | 0.62 | 0.33 | 0.00 | 0.00 |
| Euclidean Distance | 0.78 | 0.80 | 0.82 | 0.56 | 0.31 |
| Manhattan Distance | 0.65 | 0.78 | 0.73 | 0.80 | 0.73 |
| Cosine Similarity | 0.77 | 0.81 | 0.65 | 0.00 | 0.00 |

**Table A.6:** F1-Scores with the HaluEval TestSplit of Different Comparison Methods and Layer Depths for Flattened Aggregation (Non-Linear)

### A.3.2 Model Selection Statistics

| Layer Depth | Mean | Median | Std Dev | Max | Min |
|---|---|---|---|---|---|
| 1 | 0.75 | 0.77 | 0.05 | 0.81 | 0.63 |
| 2 | 0.79 | 0.80 | 0.04 | 0.82 | 0.62 |
| 3 | 0.64 | 0.72 | 0.16 | 0.82 | 0.33 |
| 4 | 0.33 | 0.28 | 0.31 | 0.81 | 0.00 |
| 5 | 0.22 | 0.04 | 0.29 | 0.81 | 0.00 |

**Table A.7:** Statistics for all Layer Depths combining the F1-Scores of all Model Selection Tables

| Aggregation Module | Mean | Median | Std Dev | Max | Min |
|---|---|---|---|---|---|
| Shared Classifier Ensemble (Linear) | 0.26 | 0.24 | 0.24 | 0.72 | 0.00 |
| Shared Classifier Ensemble (Non-Linear) | 0.36 | 0.32 | 0.25 | 0.80 | 0.00 |
| Flattened Aggregation (Linear) | 0.52 | 0.68 | 0.31 | 0.82 | 0.00 |
| Flattened Aggregation (Non-Linear) | 0.57 | 0.76 | 0.30 | 0.82 | 0.00 |

**Table A.8:** Statistics of Classifiers combining the F1-Scores of all Model Selection Tables

| Comparison Method | Mean | Median | Std Dev | Max | Min |
|---|---|---|---|---|---|
| No Comparison | 0.36 | 0.35 | 0.32 | 0.80 | 0.00 |
| Dot Product | 0.48 | 0.56 | 0.32 | 0.81 | 0.00 |
| Euclidean Norm | 0.56 | 0.72 | 0.28 | 0.80 | 0.08 |
| Manhattan Norm | 0.62 | 0.72 | 0.21 | 0.81 | 0.16 |
| Pairwise Dot Product | 0.27 | 0.24 | 0.29 | 0.81 | 0.00 |
| Euclidean Distance | 0.57 | 0.65 | 0.24 | 0.82 | 0.08 |
| Manhattan Distance | 0.54 | 0.65 | 0.26 | 0.81 | 0.00 |
| Cosine Similarity | 0.37 | 0.44 | 0.33 | 0.82 | 0.00 |

**Table A.9:** Statistics for Comparison Methods combining the F1-Scores of all Model Selection Tables

| Aggregation Module | Mean | Median | Std Dev | Max | Min |
|---|---|---|---|---|---|
| Shared Classifier Ensemble (Linear) | 0.50 | 0.49 | 0.16 | 0.72 | 0.16 |
| Shared Classifier Ensemble (Non-Linear) | 0.55 | 0.60 | 0.15 | 0.72 | 0.29 |
| Flattened Aggregation (Linear) | 0.76 | 0.78 | 0.06 | 0.81 | 0.62 |
| Flattened Aggregation (Non-Linear) | 0.78 | 0.79 | 0.05 | 0.82 | 0.63 |

**Table A.10:** Statistics of Classifiers combining the F1-Scores only of Layer 1 and 2

| Layer Depth | Mean | Median | Std Dev | Max | Min |
|---|---|---|---|---|---|
| 1 | 0.75 | 0.77 | 0.05 | 0.81 | 0.63 |
| 2 | 0.80 | 0.80 | 0.01 | 0.82 | 0.77 |

**Table A.11:** Statistics for Layer Depths 1 and 2 combining the F1-Scoress of the Flattened Aggregation (Non-Linear) Table

## A.3.3 Cross-Benchmark Evaluation

| Trained on | Tested on | | | Filtered Out |
|---|---|---|---|---|
| | **ReFact** | **HaluEval** | **TruthfulQA** | |
| ReFact | -3.01 | -31.57 | -25.73 | 9 |
| HaluEval | -3.10 | +19.60 | -11.34 | 0 |
| TruthfulQA | +1.42 | -17.89 | +12.75 | 0 |

**Table A.12:** Relative change in accuracy points when combining the cosine method with the fake fact detection results of the development LLM

| Trained on | Tested on | | | Filtered Out |
|---|---|---|---|---|
| | **ReFact** | **HaluEval** | **TruthfulQA** | |
| ReFact | -3.01 | -31.57 | -26.83 | 9 |
| HaluEval | -3.43 | +19.84 | -13.46 | 0 |
| TruthfulQA | +0.48 | -15.51 | +13.17 | 0 |

**Table A.13:** Relative change in accuracy points when combining the no comparison method with the fake fact detection results of the development LLM

## A.3.4 Pretraining / Freezing

| Selected Model | Selected Benchmark | Experiment | | |
|---|---|---|---|---|
| | | Baseline | Pretraining | Pretraining + Freezing |
| Both | All | -0.51 | 0.32 | 0.25 |
| | Final Training | 15.54 | 13.39 | 13.35 |
| | Non-Final Training | -8.53 | -6.22 | -6.30 |
| No Comparison | All | -0.39, | -1.12 | -0.53 |
| | Final Training | 15.69 | 13.72 | 14.57 |
| | Non-Final Training | -8.43 | -8.53 | -8.08 |
| Cosine | All | -0.73 | 1.58 | 0.85 |
| | Final Training | 15.26 | 13.10 | 12.38 |
| | Non-Final Training | -8.72 | -4.18 | -4.91 |

**Table A.14:** Relative fake fact detection improvement regarding the effect of pretraining and the weight freezing of the aggregation module compared to the Baseline setting of section subsection 5.3.1 calculated on the test splits of different benchmarks averaged over all executed runs

| Trained on | Finetuned on | Tested on | | | Filtered Out |
|---|---|---|---|---|---|
| | | ReFact | HaluEval | TruthfulQA | |
| ReFact | HaluEval | 0.53 | 0.80 | 0.42 | 0 |
| ReFact | TruthfulQA | 0.28 | 0.24 | 0.51 | 0 |
| HaluEval | ReFact | 0.39 | 0.46 | 0.30 | 0 |
| HaluEval | TruthfulQA | 0.39 | 0.46 | 0.70 | 0 |
| TruthfulQA | ReFact | 0.41 | 0.26 | 0.25 | 7 |
| TruthfulQA | HaluEval | 0.52 | 0.79 | 0.43 | 0 |

**Table A.15:** F1 scores for different benchmarks using the cosine model with pretraining

| Trained on | Finetuned on | Tested on | | | Filtered Out |
|---|---|---|---|---|---|
| | | ReFact | HaluEval | TruthfulQA | |
| ReFact | HaluEval | 0.56 | 0.79 | 0.42 | 0 |
| ReFact | TruthfulQA | 0.43 | 0.39 | 0.63 | 0 |
| HaluEval | ReFact | 0.51 | 0.46 | 0.36 | 5 |
| HaluEval | TruthfulQA | 0.48 | 0.50 | 0.68 | 0 |
| TruthfulQA | ReFact | 0.65 | 0.50 | 0.54 | 8 |
| TruthfulQA | HaluEval | 0.57 | 0.80 | 0.45 | 0 |

**Table A.16:** F1 scores for different benchmarks using the no comparison model with pretraining

| Trained on | Finetuned on | Tested on | | | Filtered Out |
|---|---|---|---|---|---|
| | | ReFact | HaluEval | TruthfulQA | |
| ReFact | HaluEval | 0.53 | 0.80 | 0.42 | 0 |
| ReFact | TruthfulQA | 0.33 | 0.42 | 0.62 | 0 |
| HaluEval | ReFact | 0.38 | 0.39 | 0.25 | 0 |
| HaluEval | TruthfulQA | 0.40 | 0.50 | 0.70 | 0 |
| TruthfulQA | ReFact | 0.52 | 0.34 | 0.36 | 4 |
| TruthfulQA | HaluEval | 0.55 | 0.80 | 0.43 | 0 |

**Table A.17:** F1 scores for different benchmarks using the cosine model with pretraining and freezed aggregation module

| Trained on | Finetuned on | Tested on | | | Filtered Out |
|---|---|---|---|---|---|
| | | ReFact | HaluEval | TruthfulQA | |
| ReFact | HaluEval | 0.56 | 0.80 | 0.44 | 0 |
| ReFact | TruthfulQA | 0.49 | 0.45 | 0.67 | 0 |
| HaluEval | ReFact | 0.44 | 0.42 | 0.38 | 6 |
| HaluEval | TruthfulQA | 0.47 | 0.50 | 0.67 | 0 |
| TruthfulQA | ReFact | 0.0 | 0.0 | 0.0 | 10 |
| TruthfulQA | HaluEval | 0.54 | 0.77 | 0.39 | 0 |

**Table A.18:** F1 scores for different benchmarks using the no comparison model with pretraining and freezed aggregation module

| Trained on | Finetuned on | Tested on | | | Filtered Out |
|---|---|---|---|---|---|
| | | ReFact | HaluEval | TruthfulQA | |
| ReFact | HaluEval | -2.80 | 20.10 | -10.82 | 0 |
| ReFact | TruthfulQA | -0.18 | -10.09 | 9.77 | 0 |
| HaluEval | ReFact | 3.19 | -0.17 | 0.28 | 0 |
| HaluEval | TruthfulQA | 0.72 | -5.03 | 15.68 | 0 |
| TruthfulQA | ReFact | 1.81 | -4.56 | 0.67 | 7 |
| TruthfulQA | HaluEval | -3.67 | 20.16 | -11.34 | 0 |

**Table A.19:** Relative change in accuracy points for the fake fact detection results using the development LLM by combining the cosine classifier with pretraining

| Trained on | Finetuned on | Tested on | | | Filtered Out |
|---|---|---|---|---|---|
| | | ReFact | HaluEval | TruthfulQA | |
| ReFact | HaluEval | -2.74 | 19.27 | -14.76 | 0 |
| ReFact | TruthfulQA | 1.17 | -18.15 | 11.63 | 0 |
| HaluEval | ReFact | 2.23 | -8.98 | -12.77 | 5 |
| HaluEval | TruthfulQA | 2.08 | -9.96 | 13.34 | 0 |
| TruthfulQA | ReFact | -3.01 | -31.57 | -26.83 | 8 |
| TruthfulQA | HaluEval | -2.08 | 19.73 | -13.21 | 0 |

**Table A.20:** Relative change in accuracy points for the fake fact detection results using the development LLM by combining the no comparison classifier with pretraining and freezed aggregation module

| Trained on | Finetuned on | Tested on | | | Filtered Out |
|---|---|---|---|---|---|
| | | ReFact | HaluEval | TruthfulQA | |
| ReFact | HaluEval | -3.86 | 19.96 | -10.98 | 0 |
| ReFact | TruthfulQA | 0.45 | -18.03 | 11.38 | 0 |
| HaluEval | ReFact | 2.98 | 3.31 | 0.20 | 0 |
| HaluEval | TruthfulQA | 0.90 | -7.23 | 14.31 | 0 |
| TruthfulQA | ReFact | 0.65 | -8.28 | -0.74 | 4 |
| TruthfulQA | HaluEval | -2.86 | 20.32 | -11.50 | 0 |

**Table A.21:** Relative change in accuracy points for the fake fact detection results using the development LLM by combining the cosine classifier with pretraining and freezed aggregation module

| Trained on | Finetuned on | Tested on | | | Filtered Out |
|---|---|---|---|---|---|
| | | ReFact | HaluEval | TruthfulQA | |
| ReFact | HaluEval | -2.53 | 19.03 | -14.88 | 0 |
| ReFact | TruthfulQA | 1.75 | -24.63 | 11.63 | 0 |
| HaluEval | ReFact | 3.01 | -11.86 | -10.37 | 6 |
| HaluEval | TruthfulQA | 1.45 | -6.58 | 13.42 | 0 |
| TruthfulQA | ReFact | 0.0 | 0.0 | 0.0 | 10 |
| TruthfulQA | HaluEval | -2.62 | 18.84 | -14.15 | 0 |

**Table A.22:** Relative change in accuracy points for the fake fact detection results using the development LLM by combining the no comparison classifier with pretraining and freezed aggregation module

### A.3.5 Baseline Comparison

| Selected Model | | Selected Benchmark | | |
|---|---|---|---|---|
| | | All | Final Training | Non-Final Training |
| Own Method | Both | -0.51 | 15.54 | -8.53 |
| | No Comparison | -0.39 | 15.69 | -8.43 |
| | Cosine | -0.73 | 15.26 | -8.72 |
| Baselines | All | -2.03 | 12.15 | -9.12 |
| | LastLayer | -2.24 | 13.82 | -10.27 |
| | MiddleLayer | -0.86 | 12.79 | -7.68 |
| | Stacked Layers | -3.72 | 8.07 | -9.62 |
| | AllLayersEnsemble | -1.29 | 13.87 | -8.87 |

**Table A.23:** Relative fake fact detection improvement comparing the baselines and proposed methods in the benchmark shift setting of subsection 5.3.1 calculated on the test splits of different benchmarks over all executed runs

| Trained on | Tested on | | | Filtered Out |
|---|---|---|---|---|
| | ReFact | HaluEval | TruthfulQA | |
| ReFact | 0.65 | 0.50 | 0.54 | 9 |
| HaluEval | 0.54 | 0.78 | 0.43 | 0 |
| TruthfulQA | 0.40 | 0.42 | 0.59 | 0 |

**Table A.24:** F1 scores for different benchmarks with Last Layer without pretraining

| Trained on | Tested on | | | Filtered Out |
|---|---|---|---|---|
| | ReFact | HaluEval | TruthfulQA | |
| ReFact | 0.0 | 0.0 | 0.0 | 10 |
| HaluEval | 0.59 | 0.77 | 0.42 | 0 |
| TruthfulQA | 0.38 | 0.41 | 0.55 | 0 |

**Table A.25:** F1 scores for different benchmarks with baseline Middle Layer without pretraining

| Trained on | Tested on | | | Filtered Out |
|---|---|---|---|---|
| | ReFact | HaluEval | TruthfulQA | |
| ReFact | 0.65 | 0.50 | 0.53 | 8 |
| HaluEval | 0.52 | 0.76 | 0.42 | 0 |
| TruthfulQA | 0.36 | 0.35 | 0.43 | 0 |

**Table A.26:** F1 scores for different benchmarks with baseline Stacked Layers without pretraining

| Trained on | Tested on | | | Filtered Out |
|---|---|---|---|---|
| | ReFact | HaluEval | TruthfulQA | |
| ReFact | 0.39 | 0.13 | 0.03 | 7 |
| HaluEval | 0.52 | 0.79 | 0.41 | 0 |
| TruthfulQA | 0.47 | 0.47 | 0.64 | 0 |

**Table A.27:** F1 scores for different benchmarks with baseline All Layers Ensemble

### A.3.6 Sequential Labeling

| Model | Tagging Scheme | | |
|---|---|---|---|
| | IO | BIO | BIOES |
| No Comparison | 0.10 | 0.02 | 0.02 |
| Cosine Similarity | 0.00 | 0.00 | 0.00 |
| Last Layer | 0.00 | 0.00 | 0.00 |
| Middle Layer | 0.00 | 0.11 | 0.00 |
| Stacked Layers | 0.00 | 0.11 | 0.00 |
| All Layers Ensemble | 0.01 | 0.15 | 0.04 |

**Table A.28:** F1 Scores for the Localization Experiment without CRF

| Model | Tagging Scheme | Accuracy | Precision | Recall | F1 Score |
|---|---|---|---|---|---|
| No Comparison | IO | 0.96 | 0.60 | 0.13 | 0.22 |
| No Comparison | BIO | 0.95 | 0.47 | 0.27 | 0.35 |
| No Comparison | BIOES | 0.95 | 0.57 | 0.15 | 0.24 |
| Cosine Similarity | IO | 0.96 | 0.59 | 0.14 | 0.23 |
| Cosine Similarity | BIO | 0.95 | 0.50 | 0.26 | 0.34 |
| Cosine Similarity | BIOES | 0.95 | 0.52 | 0.23 | 0.32 |
| Last Layer | IO | 0.95 | 0.28 | 0.13 | 0.18 |
| Last Layer | BIO | 0.95 | 0.48 | 0.09 | 0.16 |
| Last Layer | BIOES | 0.95 | 0.38 | 0.04 | 0.07 |
| Middle Layer | IO | 0.96 | 0.51 | 0.13 | 0.20 |
| Middle Layer | BIO | 0.95 | 0.41 | 0.20 | 0.27 |
| Middle Layer | BIOES | 0.95 | 0.52 | 0.14 | 0.22 |
| Stacked Layers | IO | 0.96 | 0.00 | 0.00 | 0.00 |
| Stacked Layers | BIO | 0.95 | 0.51 | 0.23 | 0.32 |
| Stacked Layers | BIOES | 0.95 | 0.59 | 0.08 | 0.14 |
| All Layers Ensemble | IO | 0.96 | 0.54 | 0.17 | 0.26 |
| All Layers Ensemble | BIO | 0.95 | 0.53 | 0.13 | 0.21 |
| All Layers Ensemble | BIOES | 0.95 | 0.48 | 0.18 | 0.26 |

**Table A.29:** Other Metrics for the Localization Experiment with CRF

| Model | Tagging Scheme | Accuracy | Precision | Recall | F1 Score |
|---|---|---|---|---|---|
| No Comparison | IO | 0.96 | 0.75 | 0.06 | 0.10 |
| No Comparison | BIO | 0.95 | 0.86 | 0.01 | 0.02 |
| No Comparison | BIOES | 0.95 | 0.82 | 0.01 | 0.02 |
| Cosine Similarity | IO | 0.96 | 0.00 | 0.00 | 0.00 |
| Cosine Similarity | BIO | 0.95 | 1.00 | 0.00 | 0.00 |
| Cosine Similarity | BIOES | 0.95 | 1.00 | 0.00 | 0.00 |
| Last Layer | IO | 0.96 | 0.00 | 0.00 | 0.00 |
| Last Layer | BIO | 0.95 | 0.00 | 0.00 | 0.00 |
| Last Layer | BIOES | 0.95 | 0.00 | 0.00 | 0.00 |
| Middle Layer | IO | 0.96 | 0.00 | 0.00 | 0.00 |
| Middle Layer | BIO | 0.95 | 0.50 | 0.06 | 0.11 |
| Middle Layer | BIOES | 0.95 | 0.00 | 0.00 | 0.00 |
| Stacked Layers | IO | 0.96 | 0.00 | 0.00 | 0.00 |
| Stacked Layers | BIO | 0.95 | 0.50 | 0.06 | 0.11 |
| Stacked Layers | BIOES | 0.95 | 0.00 | 0.00 | 0.00 |
| All Layers Ensemble | IO | 0.96 | 1.00 | 0.01 | 0.01 |
| All Layers Ensemble | BIO | 0.95 | 0.55 | 0.09 | 0.15 |
| All Layers Ensemble | BIOES | 0.95 | 0.86 | 0.02 | 0.04 |

**Table A.30:** Other Metrics for the Localization Experiment without CRF